\documentclass[journal ]{new-aiaa}
\usepackage[utf8]{inputenc}
\usepackage{textcomp}

\usepackage{graphicx}
\usepackage{amsmath}
\usepackage[version=4]{mhchem}
\usepackage{siunitx}
\usepackage{longtable,tabularx}
\usepackage{siunitx}
\usepackage[utf8]{inputenc}
\usepackage{subcaption}
\usepackage{booktabs} 
\usepackage{longtable,tabularx}
\usepackage{algorithmic}
\usepackage{booktabs}
\usepackage{subcaption}
\usepackage{cleveref}
\usepackage{cancel}
\usepackage[dvipsnames]{xcolor}
\usepackage[normalem]{ulem} 

\title{Investigation of risk-aware MDP and POMDP contingency management autonomy for UAS}

\author{Prashin Sharma \footnote{AI Researcher, Boeing Research and Technology. This work was performed when author was a Ph.D. candidate at University of Michigan, Ann Arbor, Member AIAA}}
\affil{University of Michigan, 2505 Hayward St., Ann Arbor, MI 48109}

\author{Benjamin Kraske \footnote{Ph.D. Student, Aerospace Engineering Sciences, University of Colorado Boulder, Student Member AIAA}} 
\affil{University of Colorado Boulder, 3775 Discovery Dr, Boulder, CO 80303}

\author{Joseph Kim  \footnote{Ph.D. Candidate, Robotics Department, University of Michigan, Ann Arbor, Student Member AIAA}}
\affil{University of Michigan, 2505 Hayward St., Ann Arbor, MI 48109}

\author{Zakariya  Laouar \footnote{Ph.D. Student, Aerospace Engineering Sciences, University of Colorado Boulder}, Zachary Sunberg \footnote{Assistant Professor, Aerospace Engineering Sciences, University of Colorado Boulder, Member AIAA}}
\affil{University of Colorado Boulder, 3775 Discovery Dr, Boulder, CO 80303}

\author{ Ella Atkins \footnote{Fred D. Durham Professor and Head, Kevin T. Crofton Aerospace and Ocean Engineering Department, Virginia Tech, AIAA Fellow} }
\affil{Virginia Tech, Randolph Hall, 460 Old Turner St., Blacksburg, VA 24061}

\begin{document}

\maketitle

\begin{abstract}

Unmanned aircraft systems (UAS) are being increasingly adopted for a variety of applications. The risk UAS poses to people and property must be kept to acceptable levels. This paper proposes risk-aware contingency management autonomy to prevent an accident in the event of component malfunction, specifically propulsion unit failure and/or battery degradation. The proposed autonomy is modeled as a Markov Decision Process (MDP) whose solution is a contingency management policy that appropriately executes emergency landing, flight termination or continuation of planned flight actions. Motivated by the potential for errors in fault/failure indicators, partial observability of the MDP state space is investigated. The performance of optimal policies is analyzed over varying observability conditions in a high fidelity simulator. Results indicate that both partially observable MDP (POMDP) and maximum a posteriori MDP policies had similar performance over different state observability criteria given the nearly deterministic state transition model.        

\end{abstract}

\section{Introduction}
Multicopter unmanned aircraft systems (UAS) are becoming popular for operations such as inspection, surveillance, and package delivery. Urban UAS operation at low altitude can expose an overflown population to nontrivial risk due to uncertainty in actuator and battery performance, external disturbances, potential for lost link, and, to-date, a void in UAS community standards related to system redundancy or resilience.  A survey conducted of 1500 UAS or ``drone" companies \cite{osborne2019uas} shows that UAS have a relatively high failure rate of $10^{-3}$ per flight hour. Multicopter battery degradation and motor failures are frequent contributing factors to UAS accidents because of their limited redundancy and low operational margins to achieve low operational cost. 

A high UAS component failure rate can be addressed by adding redundant systems with appropriate switching logic to manage them. However, UAS thrust, weight, and cost constraints discourage the safety-critical system triple redundancy required in commercial transport aviation. Although current UAS autopilots can build and accurately follow nominal flight plans, they are not resilient to most system failures and harsh environmental conditions, e.g., precipitation, strong wind gusts or shear. Automated emergency landing planning and contingency management are required to improve UAS operational safety by making real-time flight planning as well as guidance, navigation, and control (GNC) decisions. Such contingency management autonomy will enable an UAS to avoid collision and land at a safe unpopulated site rather than descending uncontrolled and/or unpowered (e.g., with a parachute) into whatever lies below the UAS flight path.   

In this paper a risk aware Contingency Management Autonomy (CMA) formulated as a Markov Decision Process (MDP) is proposed. CMA is aware of system risk from available battery and motor prognosis information.  Although CMA is modelled as fully observable, in reality systems are partially observable. Hence a partially observable MDP (POMDP) CMA representation is also investigated in this work. While POMDPs provide a direct method of representing component state uncertainty, these formulations also require greater computational overhead to solve. As such, it is important to determine if the events under consideration necessitate representation as a POMDP or if an MDP is sufficient. Our work investigates the value of partial observability in the context of this CMA implementation.

The contributions of the paper are as follows:
\begin{itemize}
    \item Design and evaluation of risk-aware MDP and POMDP formulations for small UAS Contingency Management Autonomy (CMA).
    \item Development of an urban CMA-centric simulation framework based on experimentally validated UAS component models. 
    \item Comparison of MDP and POMDP CMA models and their performance. 
\end{itemize}

This paper is organized as follows. A literature review of UAS contingency management methods and prognosis techniques is presented in Section \ref{sec:Lit}. A problem statement in Section \ref{sec:Prob} defines the scope of CMA for UAS. MDP and POMDP CMA models are specified in Section \ref{sec:Methods}. The simulation architecture used for evaluating MDP and POMDP CMA policies is detailed in Section \ref{sec:Sim}. Results of the optimal policy implementations and a discussion are presented in Sections \ref{sec:Results} and \ref{sec:Discussion}, respectively. Conclusions are presented in Section \ref{sec:Conclusion}.

\section{Literature Review}
\label{sec:all_lit}
\subsection{Contingency Management Autonomy}
\label{sec:Lit}
A key to safe contingency management autonomy (CMA) is establishing context-appropriate optimization and evaluation metrics \cite{ochoa2022urban} that ensure the UAS remains within its potentially degraded safe operating envelope, flies well-clear of terrain, buildings, and other aircraft, and minimizes overflight risk to people and property \cite{di2017evaluating}. Prognostics methods provide a set of tools for predicting component failures. However, prognostics information alone is not helpful unless used in an active system for preventative maintenance or contingency management.  Ref. \cite{balaban2013modeling} proposes the Prognostics-based Decision Making (PDM) architecture consisting of four main elements: a diagnoser (DX), decision maker (DM), vehicle simulation (VS) and the vehicle itself. The prognostics problem is formulated as a constraint satisfaction problem (CSP) and solved using backtracking search and particle filtering. In this framework, mission waypoints are defined a-priori; waypoints are assumed reachable even in the presence of faults. A similar prognostics architecture is proposed in \cite{balaban2013development} and implemented on an unmanned ground vehicle. In \cite{schacht2019} the authors proposed a multicopter mission planning strategy that incorporates battery State of Charge (SoC) and State of Health (SoH) to generate updated mission plans. The planning problem is formulated as an optimization problem to minimize total energy consumed by the multicopter subject to nonlinear constraints defined by UAS, brushless motor, and battery dynamics.  In reference \cite{tang2008} and follow-on work \cite{tang2010} the authors present an Automated Contingency Planner enhanced by prognostic information. Online optimization determines a minimum cost reconfiguration for the system and components. A receding horizon planner is utilized in \cite{zhang2014} to incorporate the constraints determined from prognostics information. 

A survey on commercial UAS safety and reliability was conducted by \cite{osborne2019uas} showing that the battery system was the third most likely critical subsystem to fail. Contingency management for battery systems starts with prognostics and health management (PHM), and several researchers have studied precise estimation of battery remaining useful life (RUL) computed from battery state of charge (SoC). Such methods have used Extended Kalman Filter (EKF) \cite{schacht2018prognosis}, Unscented Kalman Filter (UKF) \cite{he2013state}, unscented transform \cite{daigle2010improving}, particle filter \cite{dalal2011lithium}, neural network \cite{obeid2020supervised}, and Gaussian Process Regression (GPR) \cite{wu2016review, liu2013prognostics} formulations.  Sharma et al. \cite{sharma2021prognostics} proposed multi-battery reconfiguration for UAS using a prognostics-informed Markov Decision Process (MDP). An optimal battery switching MDP policy for a UAS with two battery packs to respond appropriately to observed battery pack degradation and remaining flight time. Case studies examined optimal battery reconfiguration performance over several UAS mission scenarios. 

Another critical subsystem failure is the motor and propeller damage/failure, which may directly trigger multicopter instability. The characterization and prognosis of such damage were studied by Brown et al. \cite{brown2015characterization} using measurable failure modes. The damage/failure was indirectly detected from anomalies in measurable parameters such as control signal, angular velocity, etc. For example, monitoring motor control signal data from the autopilot indirectly observed the damage in a specific motor, as the remaining rotors compensate for the failure. Such self-monitoring failure modes contributed to structural health monitoring (SHM) and damage prognosis (DP) used in UAS. Once the rotor failure is detected, fault-tolerant control (FTC) can re-allocate controllers to maintain safe flight. Zhang et al \cite{zhang2013development} provides a detailed overview of existing fault detection and diagnosis (FDD) and FTC in unmanned rotorcraft systems. Particularly, Ref. \cite{dydek2012adaptive} combined/ composite model reference adaptive controller (CMRAC) to safely maintain and land a quadrotor with single thruster failure. Loss of two or more thrusters in a quadrotor has also been studied as FTC in \cite{mueller2014stability}, though the resulting reduced attitude kinematics control authority can only provide periodic solutions that cannot support touchdown at a specific landing site.

Kim, et al. \cite{kim2021assured} developed a systematic auto-mitigation strategy for Advanced Air Mobility (AAM), generating safe contingency actions in case of rotor failures in a multicopter. The paper offered Assured Contingency Landing Management (ACLM) sub-component logic flow and mathematical derivation of degraded controllability and landing site reachability. Case studies show contingency landing site selection based on risk-based cost metrics using offline and online flight planners, an extension of previous work in online multicopter emergency landing planning given battery energy degradation \cite{ten2017emergency}. Our paper is distinct in its computation of UAS contingency management using Probabilistic Model Checking (PMC) with Markov Decision Process (MDP) and Partially Observable Markov Decision Process (POMDP) formulations.    

\subsection{Observable versus Partially Observable Markov Decision Process Models}
\label{sec:Lit_vs}
As Sharma et al. demonstrate in \cite{sharma2021prognostics}, the MDP framework supports reasoning about aleatoric uncertainty to inform safety-critical contingency management decisions. However, in cases where the state of the system is not fully observable, reasoning about epistemic uncertainty is needed. A problem with these uncertainties can be formulated as a POMDP. Sunberg et al. showed that it can be valuable to infer the hidden states of a problem to obtain a better performing policy than a predefined baseline \cite{zsunberg_value_of_inferring}. A critical advantage of POMDPs over MDPs lies in the robustness gained from adopting a probabilistic representation of state informed through observations \cite{kurniawati2022partially}. The optimal actions under each paradigm maximize a defined reward function, yet only in POMDPs can actions fulfill the tangential goal of gathering information to improve its belief  \cite{kaelbling1998planning}. The current national airspace collision avoidance system known as ACAS-X uses a POMDP framework to provide robust surveillance and advisory logic to aviation pilots \cite{kochenderfer2012next}. 

The problem of detecting faults on an autonomous system can be difficult and typically requires inferring the faults from potentially incomplete and noisy measurements. A partially observable fault state cannot be characterized properly when the FDD/FTC problem is modeled as an MDP. A POMDP framework allows for maintaining a belief representation of the fault through a probability distribution over fault states. The authors in \cite{Goel2000, Mehra1998, Zhang1998} use a framework for maintaining a belief over fault states or modes allowing for more robust recovery after failure. 

Although POMDPs can be more expressive and robust than MDPs, exact solutions to general finite-horizon POMDPs are PSPACE-complete in the worst case leading to approximation methods \cite{kochenderfer2022algorithms}. The designer must then trade off satisfactory solutions and solution speed. The authors in \cite{littman1995learning} present QMDP which makes an assumption of full observability after the first time step to find an approximately optimal policy for an infinite-horizon POMDP. This approach can lead to poorly approximating the value of information gathering actions by naively claiming full knowledge of the state process. Many approaches have attempted to overcome the \textit{curse of dimensionality} of POMDPs while still remaining near-optimal. SARSOP \cite{kurniawati_sarsop_2008} is one such approach: Kurniawati et al. exploit the notion of \textit{optimally reachable belief spaces} to improve on computational efficiency. 

To the best of our knowledge, a comprehensive study comparing MDP vs POMDP effectiveness in contingency management settings within a high fidelity simulator has not been conducted. This work aims to take a step in addressing this gap and offer a better insight into when to model a contingency management problem as an MDP vs a POMDP.

\section{Problem Statement}
\label{sec:Prob}
The goal of this research is to develop contingency management autonomy (CMA) for UAS that utilizes system state and component prognosis information to select appropriate actions to preserve UAS safety and prevent high-risk system failures. To assess the CMA a specific mission is considered for further investigation, though the presented CMA methods could also be applied to other missions. The mission considered in this paper is a  multirotor UAS executing a package delivery mission when an in-flight propulsion unit failure, battery degradation, or sudden low battery voltage event occurs. We assume the flight plan and emergency landing site coordinates are known. Multiple flight plans with varying safety margin on available battery energy are considered. Unknown information includes time of propulsion unit and/or battery failure as well as wind gust strength. 

\begin{figure*}[t!]
     \centering
     \begin{subfigure}[t]{0.47\textwidth}
         \centering
         \includegraphics[scale = 0.45]{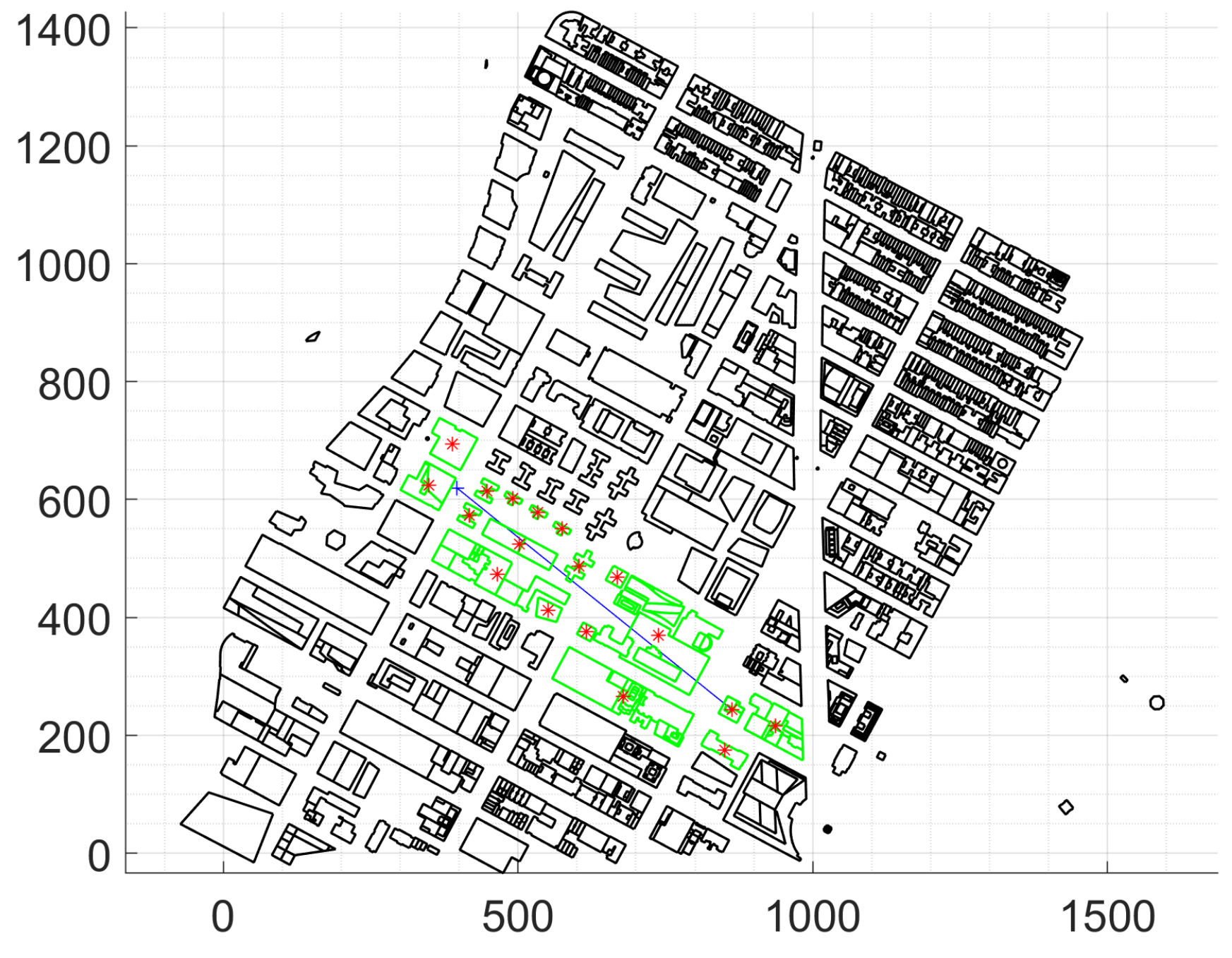}
         \caption{Top-down view of Manhattan with nominal flight path in blue line, contingency landing sites in red stars}
         \label{fig:top_down_mahattan}
     \end{subfigure}
     \hfill
     \begin{subfigure}[t]{0.47\textwidth}
         \centering
         \centering
         \includegraphics[width=3.25in, height = 2.25in]{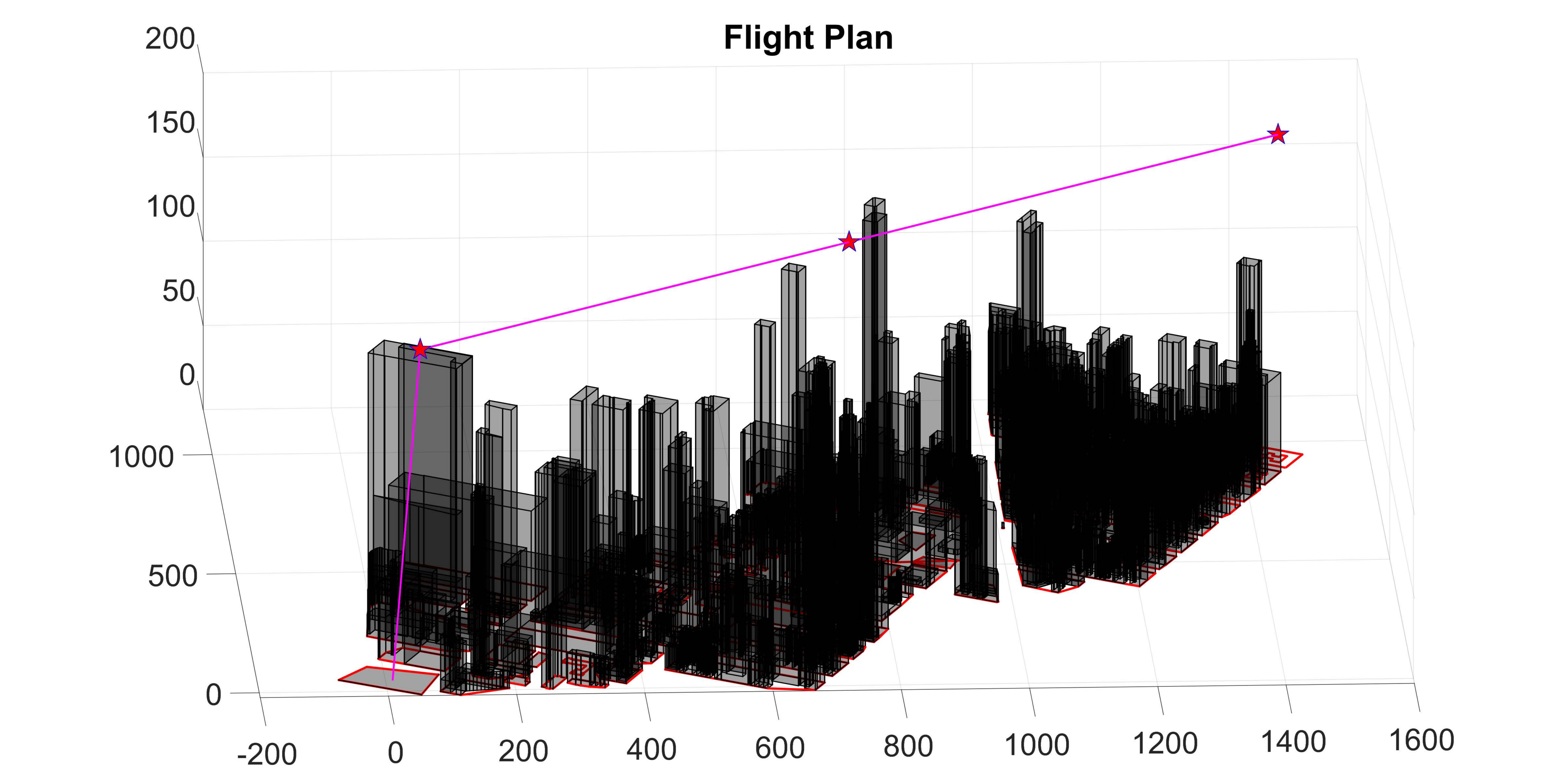}
         \caption{3-D Manhattan environment with nominal flight path in magenta line and pre-planned checkpoints in Red stars.}
         \label{fig:mahattan_3D_flightTraj}
     \end{subfigure}
     \caption{Example small UAS package delivery mission used for MDP and POMDP CMA simulations. }
 \end{figure*}

A simulation environment to assess contingency planning models was constructed using post-processed OpenStreetMap (OSM) building data from \cite{kim2022airspace}. This map data builds 3-D building structures in Southern Manhattan defined by building height, type, and outline coordinates. Emergency landing sites are found using the modified geofence polygon extraction algorithm in \cite{kim2022airspace}. Once the start and destination of the UAS flight are determined, the algorithm generates a region of interest, and potential landing sites are found using computational geometry. Our CMA is simulated with this emergency landing site map using a hexacopter that experiences a sudden in-flight fault. Figure \ref{fig:top_down_mahattan} shows a top-down view of an example package delivery mission in southern Manhattan with offline landing sites highlighted in green lines and red stars. Figure \ref{fig:mahattan_3D_flightTraj} shows a 3-D preflight contingency database as well as a nominal flight trajectory for a package delivery mission defined in our simulation.

\section{Contingency Planning  Models }
\label{sec:Methods}
The purpose of CMA formulation is to generate risk aware safety preserving actions for the UAS while considering any observed degradation in the multi-battery pack and motors. In this work CMA is modelled as a stationary infinite-horizon Markov Decision Process \cite{Martin2005}. From this MDP model, a POMDP model is constructed with varying state observability conditions. In the following section we provide details of the CMA MDP and POMDP formulations.

\subsection{MDP Model}
\label{sec:mdp_model}

 An MDP is defined by the 4-tuple $\langle S,A,p(s'|s,a),r(s,a) \rangle $, where $S$ is the finite set of system states, $A$ is allowable actions, $p(s'|s,a)$ is state transition probability tensor and $r(s,a)$ is the reward for executing action $a$ in state $s$. The MDP computes actions that maximize expected value for each state based on the Bellman equation. Classic algorithms such as Value Iteration or Policy Iteration can be used to determine optimal MDP policy $\pi^\star$. 

\subsubsection{State Space Description}

The CMA MDP is designed to minimize UAS risk by executing emergency landing options as needed. The MDP state space includes information related to motor and battery health, available battery energy and remaining useful life of motor, which collectively provide critical information to enable safe landing when possible. CMA MDP state is defined as $S = S_E \cup S_S $, $S_E =\{ C, T, FL, E\}$ and $S_S = \{ S= (FS, MH, MM, BH, RM) | FS \in \{N,ELASAP,ELPract\}\},$ $MH \in \{ NF,SF,JF \},$ $MM\in \{MM0, MM1\}, $ $BH \in \{G, M, P\},$ $RM \in \{ RM0, RM1 \}  \}$.


\noindent Elements of the MDP state vector are defined as follows: 
\begin{itemize}
\item \textit{Flight Status (FS)}:
    Due to the Markov assumption, the current state must contain relevant system status information.  The $FS$ state feature has three possible values:  $N$: Nominal flight, $ELASAP$: Executing emergency landing ASAP (as soon as possible), $ELPract$: Executing emergency landing when practical. If $FS =ELPract$ the action space reduces to $A_s=\{NoOp, Terminate, LandASAP \}$. if $FS=ELASAP$ the action space reduces to $A_s=\{NoOp, Terminate \}$. All actions are available with $FS=N$. The action availability based on $FS$ has been manually coded in the MDP. 
    
    \item \textit{Motor Health (MH)}
    This feature assumes the following values $MH=\{NF:No Fault,$ $ SF:Spalling Fault,$ $ JF:Jam Fault \}$. There exists various types of motor degradation, however in this research the scope is limited to spalling degradation, which is one of the critical failure conditions for a motor bearing. This fault was selected because significant literature related to prognosis for spalling faults is available \cite{qiu2020},\cite{zhang2009}. The occurrence of a spalling fault is assumed fully observable. When the spall area exceeds a threshold area value the motor fails. Once the motor ceases operation, it is detected as a jam fault by an Interacting Multiple Model (IMM) Kalman Filter as described in Section \ref{sec:Sim}. 
    
    \item \textit{Motor Margin (MM)} is defined numerically for a single motor by:
    \begin{equation}
        MM = 1- \sum_{i=1}^{n_f} w_i\frac{t_{FlightTime}}{t_{{RUL}_i}}
        \label{eq:MM}
    \end{equation}
    where $t_{FlightTime} = $ flight time (sec), $ t_{RUL} = $ motor Remaining Useful Life (RUL) (sec), $n_f =$ different types of motor faults, $\sum_{i=1}^{n_f}w_i =1 $. In this paper only a single spalling fault is considered, hence $n_f =1$. When a spalling fault is detected $t_{RUL}$ is calculated using Paris Law \cite{zhang2009}. For nominal motor health conditions $t_{RUL} \geq 10\times t_{FlightTime}$ because under nominal conditions the propulsion system would be designed with a high safety margin.  MDP discrete state feature $MM$ is assigned two logical values: $MM0:MM<0$ and $MM1:0\leq MM < 1 $. $MM<0$ indicates that either the motor has ceased or $t_{RUL}<t_{FlightTime}$, i.e. the motor will not be able to provide thrust for the total remaining flight duration. Range   $0\leq MM < 1 $ signifies the motor is either in nominal health or undergoing a spalling fault but expected to provide thrust for the remaining flight duration. If multiple motor faults are considered,  further investigation would be required for discretization of $MM$. In this paper only a single motor failure is considered. 

    \item \textit{Battery Health (BH)}
    is abstracted to three possible health conditions: $Good$, $Medium$ and $Poor$. Nominally $BH = Good$. If the battery experiences either power fade or capacity fade but not both $BH = Medium$. If the battery pack experiences both power and capacity fade $BH = Poor$. As battery degradation is a relatively slow process, we assume as a simplification in this work that $BH$ remains constant during a single flight.
    
    \item \textit{Reachability Margin ($RM$)}
    is defined by: 
    \begin{equation}
        RM = 1 - \frac{t_{FlightTime}}{t_{EOD}}
        \label{eq:RM}
    \end{equation}
    where $t_{EOD} = $ End of discharge time for  a series-parallel battery pack with both the batteries being used and is calculated as described in \cite{sharma2021prognostics} .  $t_{FlightTime}$ is the time the UAS takes to complete the executing flight plan.  $RM$ is discretized as $RM0:RM<0$ and $RM1:0\leq RM<1$. $RM0$ indicates the battery pack does not have sufficient energy to complete the executing flight plan. $RM1$ indicates the battery pack has sufficient energy to complete the flight plan.
    
    \item \textit{Complete(C)} assumes value $False (0)$ while the mission is in progress and $True (1)$ when the UAS completes its nominal or emergency flight plan.  All $s\in S$ with $C=1$ transition to E (end) an absorbing state.
    
    \item \textit{Terminated (T)} assumes value $False (0)$ until the flight termination action is executed at which time it latches to $True (1)$. All $s\in S$ with $T=1$ transition to $E$ (end), an absorbing state.    
    
    \item \textit{Failure (FL)} assumes value $False (0)$ unless its value is set to $True (1)$ indicating the mission has failed due to controller instability (measured by trajectory tracking error) or insufficient battery energy ($RM<0$). Other system failures are not considered in this work but would be necessary to model and manage with CMA in any autonomous system certification process.  All $s\in S$ with $FL=1$ transitions to $E$ (end), an absorbing state.
\end{itemize}

This state space is intentionally abstracted to aggregate important decision regions for contingency management into a minimal complexity representation. This state space abstraction results in  $|S_E| = 4$, $|S_S| = |FS|\times|MH|\times|MM|\times|BH|\times|RM| = 3\times3\times2\times3\times2=108$, totalling to $|S|=112$ states with $C, T, FL$, and $E$. 

\subsubsection{Action Space Description}
The CMA action space is designed to respond to component degradation scenarios considered in this paper while minimizing risk posed by the UAS. The CMA action set is $A=\{NoOp, Terminate,$ $LandASAP,$ $LandPract\}$ defined as follows:
\begin{itemize}
    \item \textit{No Operation (NoOp):}  With this action, the UAS continues executing the current flight plan. 
    
    \item \textit{Terminate:} This action indicates flight termination to minimize risk to people and property of a failed UAS on touchdown. The terminate action could be implemented as deployment of a parachute and cutting motor power. This action immediately renders the UAS inoperable.  While the UAS lands uncontrollably when executing the \textit{Terminate} action, the UAS poses less risk than a loss-of-control event would pose because a parachute reduces kinetic energy and zero torque will be applied to the propellers.
    \begin{figure*}[ht!]
     \centering
     \begin{subfigure}[t]{0.5\textwidth}
         \centering
         \includegraphics[scale = 0.45]{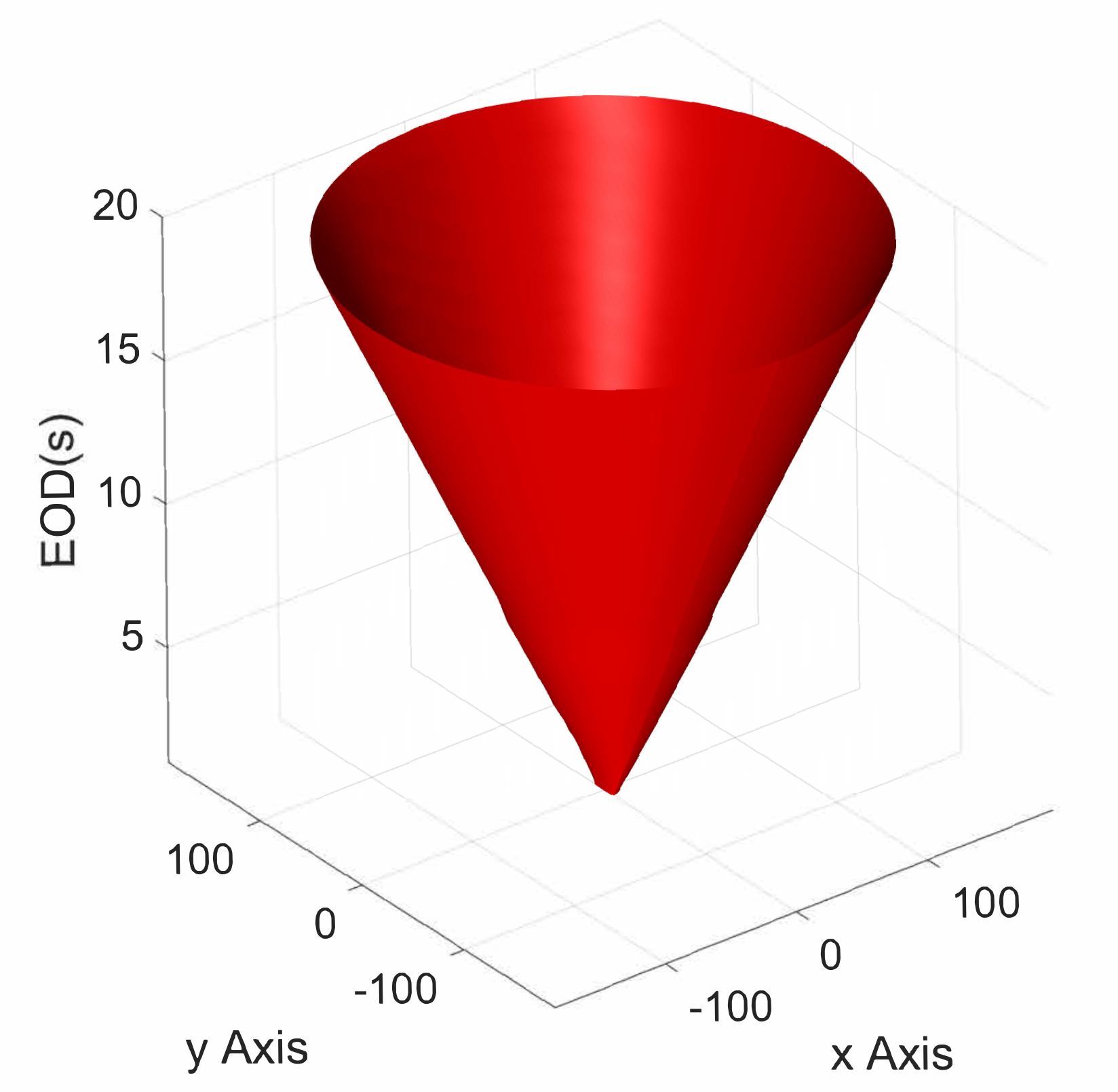}
         \label{fig:FootPrintTime}
     \end{subfigure}
     \begin{subfigure}[t]{0.4\textwidth}
         \centering
         \centering
         \includegraphics[scale = 0.4 ]{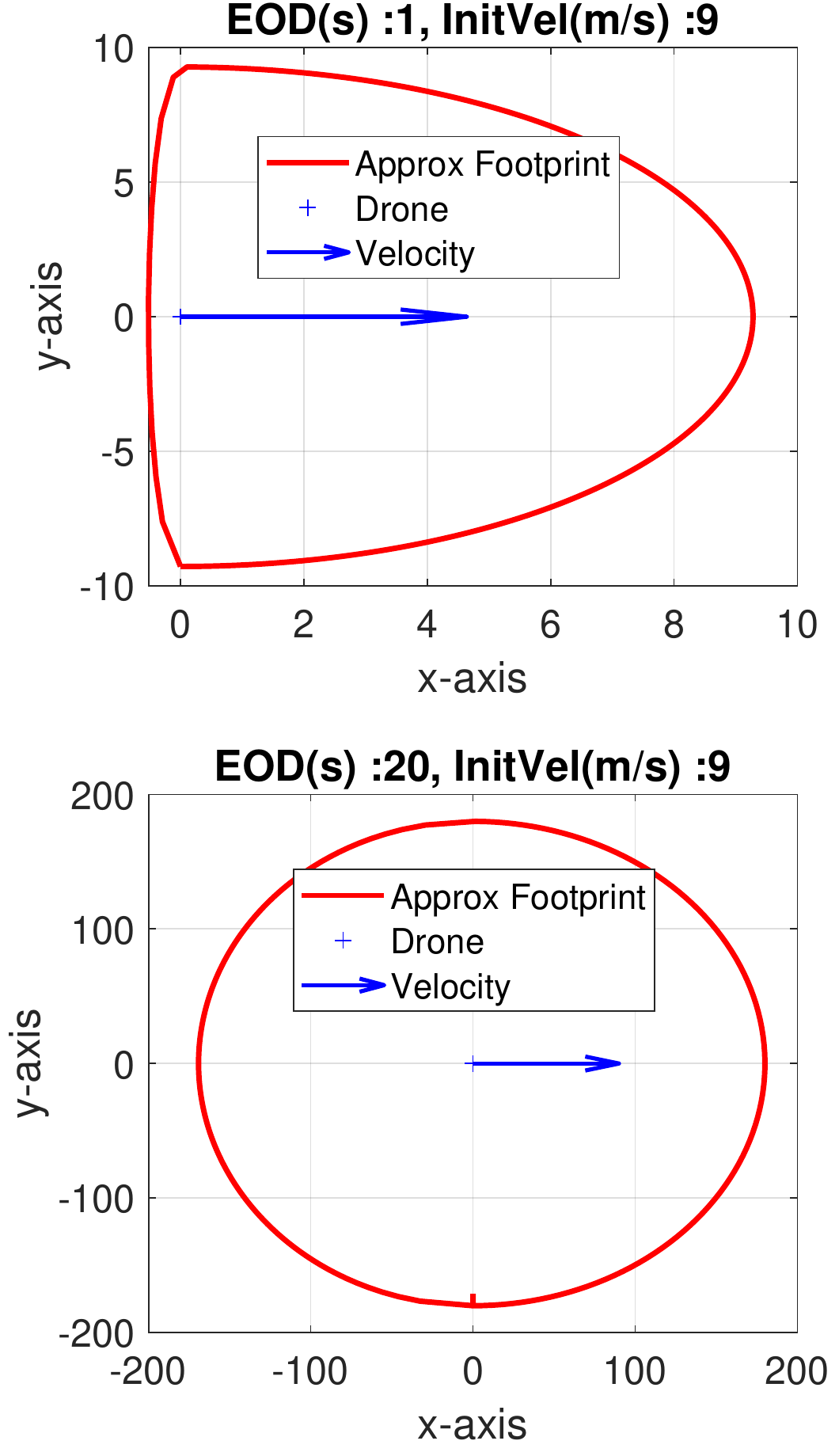}
        \label{fig:FootPrint2D}
     \end{subfigure}
      \caption{(Left) Multicopter footprint for different battery EOD values and initial velocity of $9m/s$ along the $+x$ axis. (Top Right) Footprint for an EOD value of $1sec$.  (Bottom Right) Footprint for an EOD value of $20sec$. }
      \label{fig:footprintEOD}
    \end{figure*}
    \item  \textit{LandASAP:} When this action is executed, the UAS determines a list of available emergency landing sites within its approximate reachable footprint first used for aircraft emergency landing planning in \cite{atkins2006emergency}. The approximate footprint is calculated as the maximum distance the UAS can travel by defining a minimum control effort trajectory in lateral and longitudinal directions as shown in Figure \ref{fig:footprintEOD}.  As shown multicopter footprint is a function of initial velocity and battery End of Discharge (EOD) time. The cone appearing in Figure \ref{fig:footprintEOD} shows an increase in approximate footprint with higher EOD values for a given initial velocity condition. The right half of Figure \ref{fig:footprintEOD} shows cross sections of the cone at different EOD times. As shown lower EOD values and higher initial velocities result in an asymmetric footprint cross section due to the time required for the multicopter to decelerate and reverse travel direction. $LandASAP$ then plans a minimum time trajectory to identified reachable emergency landing sites and selects the solution with maximum reachability margin $RM$. This step can be computationally expensive resulting in a preference for the $LandPract$ action defined below when possible. 
  \begin{figure}[ht!]
    \centering
    \includegraphics[width = 0.97\textwidth]{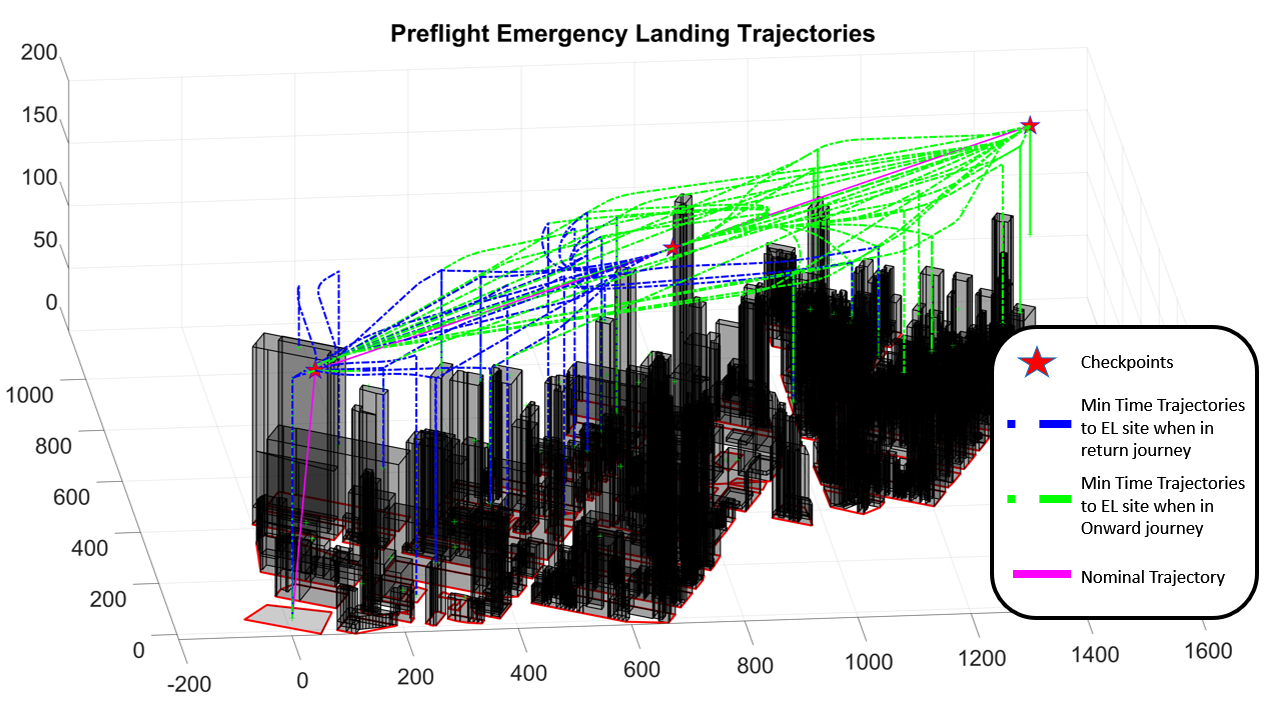}
    \caption{Emergency landing (EL) plans from checkpoints on the nominal plan computed and stored before flight.}
    \label{fig:LandPract}
\end{figure}
    \item \textit{LandPract:} With this action, the UAS utilizes landing trajectories calculated and stored before flight per \cite{kim2021assured} to the set of reachable emergency landing sites from a set of nominal flight plan checkpoints shown as $\bigstar$ in Figure \ref{fig:LandPract}. The checkpoints are selected to divide the flight plan into equal length segments also per \cite{kim2021assured}. Based on the current UAS location and its proximity to a checkpoint in the nominal flight plan, an emergency landing site from the list of pre-planned emergency landing trajectories with the maximum $RM$ is selected. An example pre-planned emergency landing trajectory is shown in Figure \ref{fig:LandPract}. This action is computationally cheaper than \textit{LandASAP} because the trajectories are selected from a database generated offline.  
\end{itemize}
    
\begin{figure}[ht!]
    \centering
    \includegraphics[width = 1.0\textwidth]{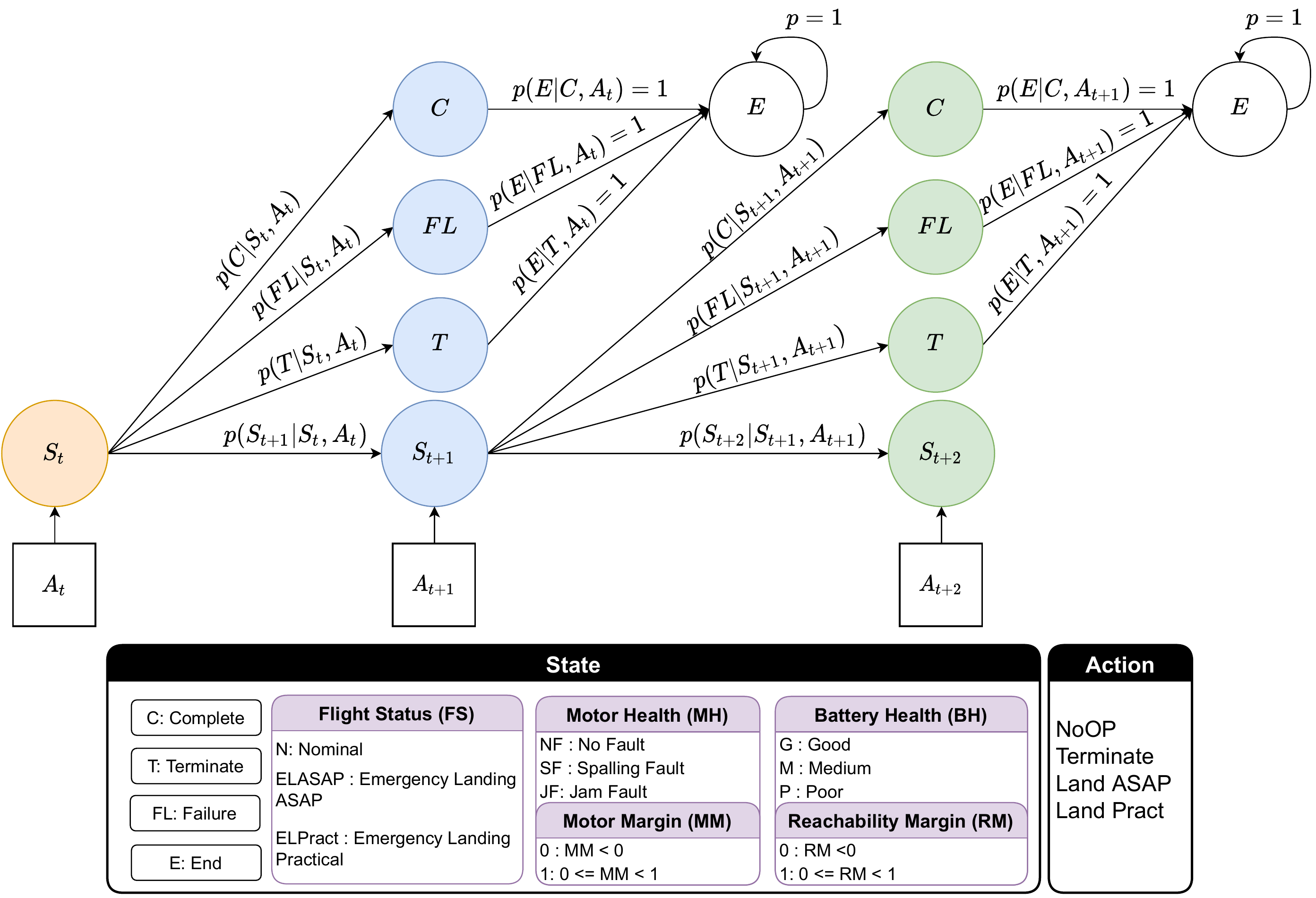}
    \caption{State transition graph for the CMA MDP.  $C: Complete$, $FL: Failure$, $T: Terminate$ represent terminal / absorbing states, $A_t : Action$ }
    \label{fig:BayesNet}
\end{figure}

\begin{figure*}[t!]
    \centering
     \begin{subfigure}[t]{0.475\textwidth}
        \centering
        \includegraphics[width=0.75\textwidth]{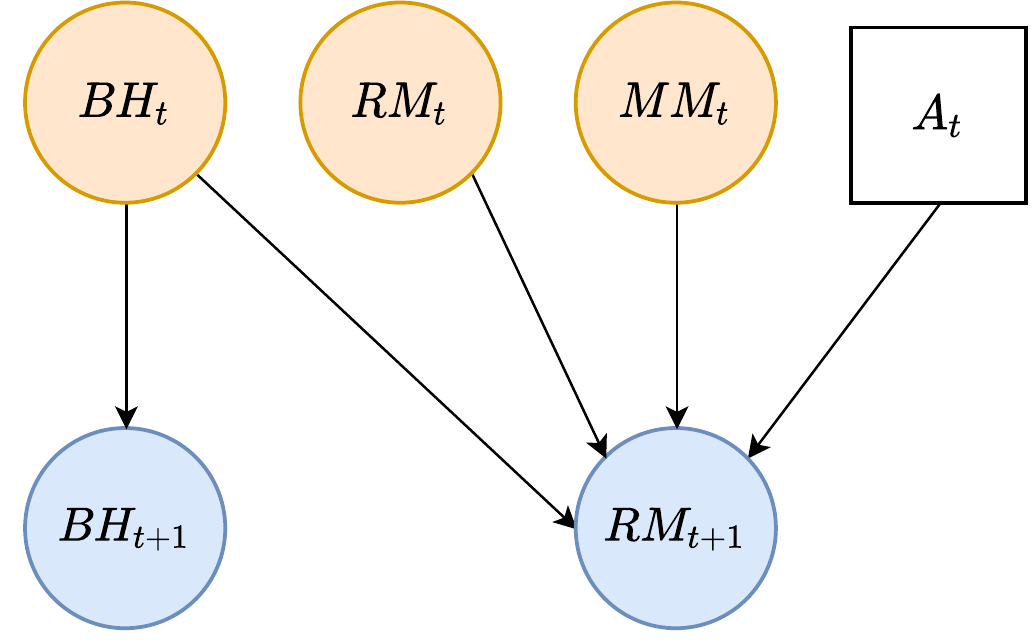}
        \caption{Battery Health and Reachability Margin}
        \label{fig:BHRM}
    \end{subfigure}
    \begin{subfigure}[t]{0.4\textwidth}
        \centering
        \includegraphics[width=0.75\textwidth]{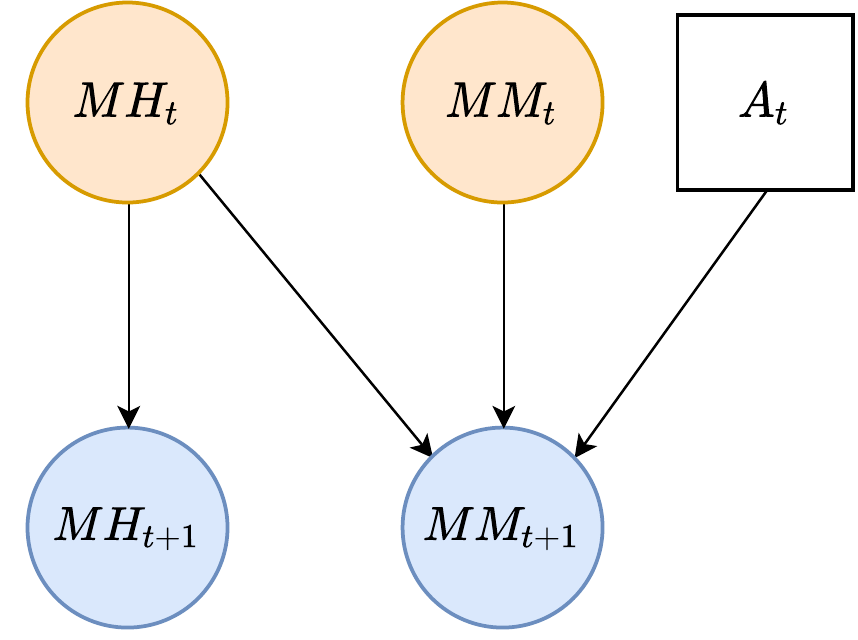}
        \caption{Motor Health and Motor Margin}
        \label{fig:MHMM}
    \end{subfigure}
    \hfill
    \begin{subfigure}[t]{0.475\textwidth}
        \centering
        \includegraphics[width=0.75\textwidth]{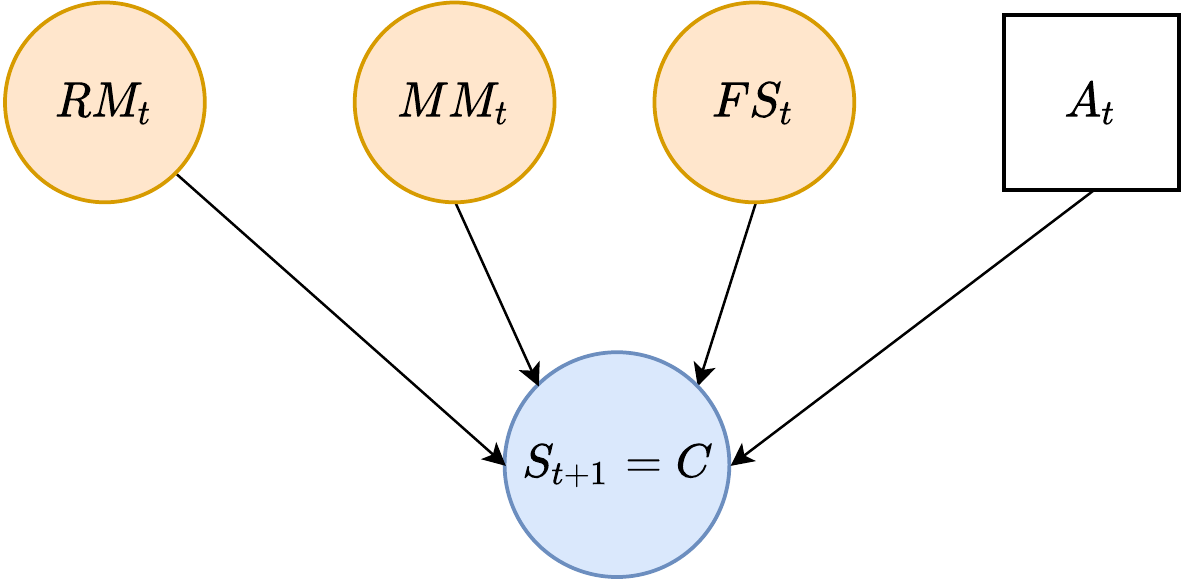}
        \caption{Completion state}
        \label{fig:C}
    \end{subfigure}
    \begin{subfigure}[t]{0.475\textwidth}
        \centering
        \includegraphics[width=0.75\textwidth]{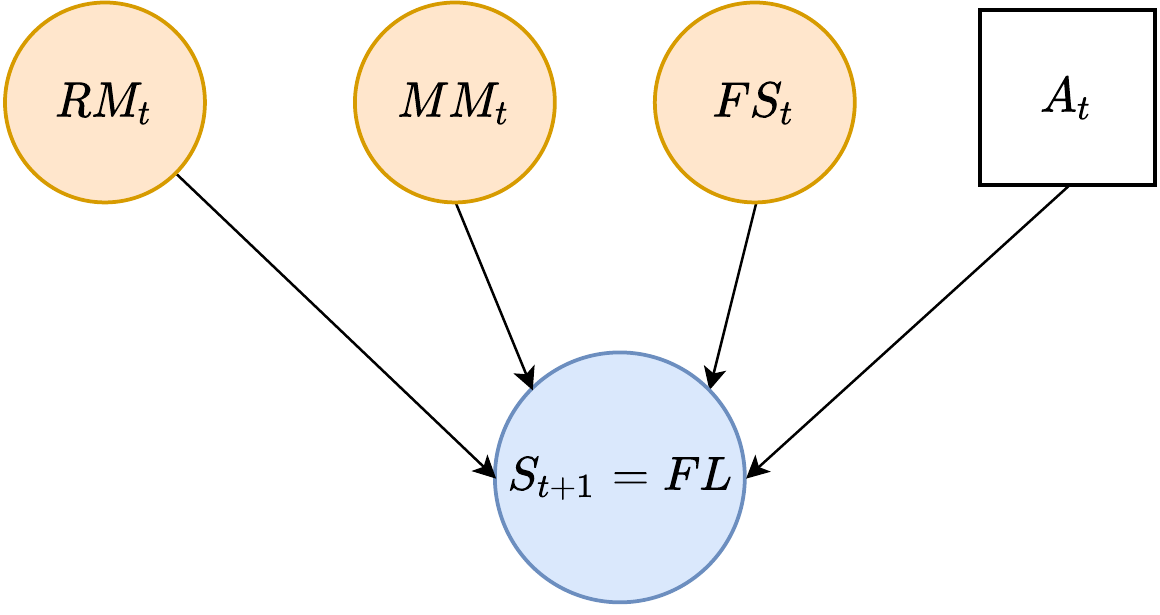}
        \caption{Failure State}
        \label{fig:FL}
    \end{subfigure}
    \hfill
    \caption{Dynamic decision networks showing dependencies of MDP state elements over time.}
    \label{fig:DecompDDN}
\end{figure*}

\subsubsection{State Transition Modelling}

State transitions for the CMA MDP are modeled as a dynamic decision network (DDN) as shown in Figure \ref{fig:BayesNet}. To simplify notation, each transient or absorbing state is labeled with the true single feature ($C$ for mission complete, $FL$ for failure, $T$ for flight terminate, $E$ end). Any other state at time $t$ is defined by feature set $S_t = \{FS,MH,MM,BH,RM\}_t$. This approach to defining the state transition probabilities rather than manually defining each element of the table enables the exploitation of conditional independence and supports explainability.
Inter-dependencies between MDP state features at time $t$ and $t+1$ are described in Figure \ref{fig:DecompDDN}.  Consider the case of motor margin ($MM_{t+1}$) at $t+1$ as shown in Figure \ref{fig:MHMM}. $MM_{t+1}$ as defined in Equation \ref{eq:MM}  from the values of $t_{RUL}$ and $t_{FlightTime}$. $t_{RUL}$ is directly influenced by motor health $MH_t$ and $t_{FlightTime}$. Hence $MM_{t+1}$ is dependent on action ($A_t$), motor margin ($MM_t$) and motor health ($MH_t$) from the previous time-step $t$. Examples of state transition probabilities are presented in Tables \ref{table:MMNoOp} and \ref{table:MH} assuming a $1Hz$ state update rate.

\begin{table}[tbh!]
 \caption{ State Transition Probabilities for $MM_{t+1}$ }
\centering
  \begin{tabular}{cccc}
    \toprule
    $MM_t$ & $MH_t$ &  $MM_{t+1}$ & $P(MM_{t+1}|MM_t,MH_t,A_t=\textit{NoOp})$ \\
      \toprule \midrule

    $MM0$	& $NF$ &	$MM0$ &		0 \\
    $MM0$	& $NF$ &	$MM1$	&	1 \\
$MM0$ &	$SF$ &	$MM0$	&	0.995 \\
$MM0$	& $SF$& $MM1$ &		0.005 \\
$MM0$ &	$JF$ &	$MM0$ &		1 \\
$MM0$ & 	$JF$ &	$MM1$ &		0 \\
$MM1$	& $NF$	& $MM0$ &	0 \\
$MM1$ &	$NF$ &	$MM1$ &		1 \\
$MM1$ &	$SF$ &	$MM0$	&	0.002809 \\
$MM1$ &	$SF$	& $MM1$ &		0.997191 \\ 
$MM1$ &	$JF$ &	$MM0$ &		1 \\
$MM1$ &	$JF$ &	$MM1$ &		0 \\
    \midrule
    \bottomrule
    \end{tabular}

\label{table:MMNoOp}
 \end{table}

\begin{table}[tbh!]
 \caption{ State Transition Probabilities for $MH_{t+1}$ }
\centering
  \begin{tabular}{cccc}
    \toprule
    $MH_t$ &  $MH_{t+1}$ & $P(MH_{t+1}|MH_t)$ \\
      \toprule \midrule
$NF$ &	$NF$	&	0.9999525 \\
$NF$ &	$SF$	&	0.0000475 \\
$NF$ &	$JF$	&	0\\
$SF$ &	$NF$	&	0\\
$SF$&	$SF$&		0.997191 \\
$SF$&	$JF$&		0.002809 \\
$JF$&	$NF$&	    0 \\
$JF$&	$SF$&		0\\
$JF$&	$JF$&		1\\
    \midrule
    \bottomrule
    \end{tabular}
\label{table:MH}
\end{table}

\subsubsection{Reward Function}
The reward function below is structured to incentivize actions which prevent failure of the UAS in degraded conditions or continue with the flight plan if possible. 
Tunable weights $w_E$, $w_S$ and $w_A$ are selected from user preference.
\begin{gather}
  R(S,A) = R(S) + R(A) \\
  R(S) = 
    \begin{cases}
     w_E(S) f_{E}(S),  &  \text{ $S \in S_E$ }\\
     w_S^T f_{S}(S),  & \text{ $S\in S_S$ } 
      \end{cases} , \; R(A) = w_A(A) f_A(A) 
\end{gather}
      
\begin{gather}
      f_{E}(S) =
      \begin{cases}
      1, & \text{$S=C$}\\
      -0.1, & \text{$S=T$}\\
      -1, & \text{$S=FL$}\\
      0, & \text{$S=E$}
      \end{cases} ,\;
       f_{A}(A) =
      \begin{cases}
      1, & \text{$A = NoOp$}\\
      -1, & \text{$A = Terminate$}\\
      -0.5, & \text{$A = LandASAP$}\\
      0.5, & \text{$A = LandPract$}
      \end{cases} 
\end{gather}

\begin{gather}
      f_{S}(S) = [f_{S_{FS}}(FS), f_{S_{MH}}(MH), f_{S_{MM}}(MM), f_{S_{BH}}(BH), f_{S_{RM}}(RM)]' \\
      f_{S_{FS}}(FS) = \begin{cases}
      1, & \text{$FS = Nominal$}\\
      -1, & \text{$FS = \{ELPract, ELASAP\} $}\\
      \end{cases},
      f_{S_{MH}} = \begin{cases}
      1, &\text{$MH = NF$}\\
      0, & \text{$MH = SF$}\\
      -1, &\text{$MH = JF$}
      \end{cases}
\end{gather}

\begin{gather}
      f_{S_{MM}}(MM) = \begin{cases}
      1, & \text{$MM = MM1$}\\
      -1, & \text{$MM = MM0 $}
      \end{cases},
      f_{S_{BH}} = \begin{cases}
      1, &\text{$BH = G$}\\
      0, & \text{$BH = M$}\\
      -1, &\text{$BH = P$}
      \end{cases}\\
      f_{S_RM}(RM) = \begin{cases}
      1, & \text{$RM = RM1$}\\
      -1, & \text{$RM = RM0 $}
      \end{cases}
\end{gather}

The rewards assigned to MDP states were chosen to favor completion of the nominal mission without interruption. However, if there is any occurrence of an off-nominal scenario such as degraded battery or motor failure, such states are penalized to encourage safe nearby landing. Similarly, action weight values are chosen to continue the mission without interruption and to prefer emergency landing actions over flight termination. The difference in values assigned for $LandASAP$ and $LandPract$ action occurs because of the higher computation cost of executing $LandASAP$ compared to $LandPract$.

\begin{figure}[ht!]
    \centering
    \includegraphics[width=6.5in, height = 2.5in]{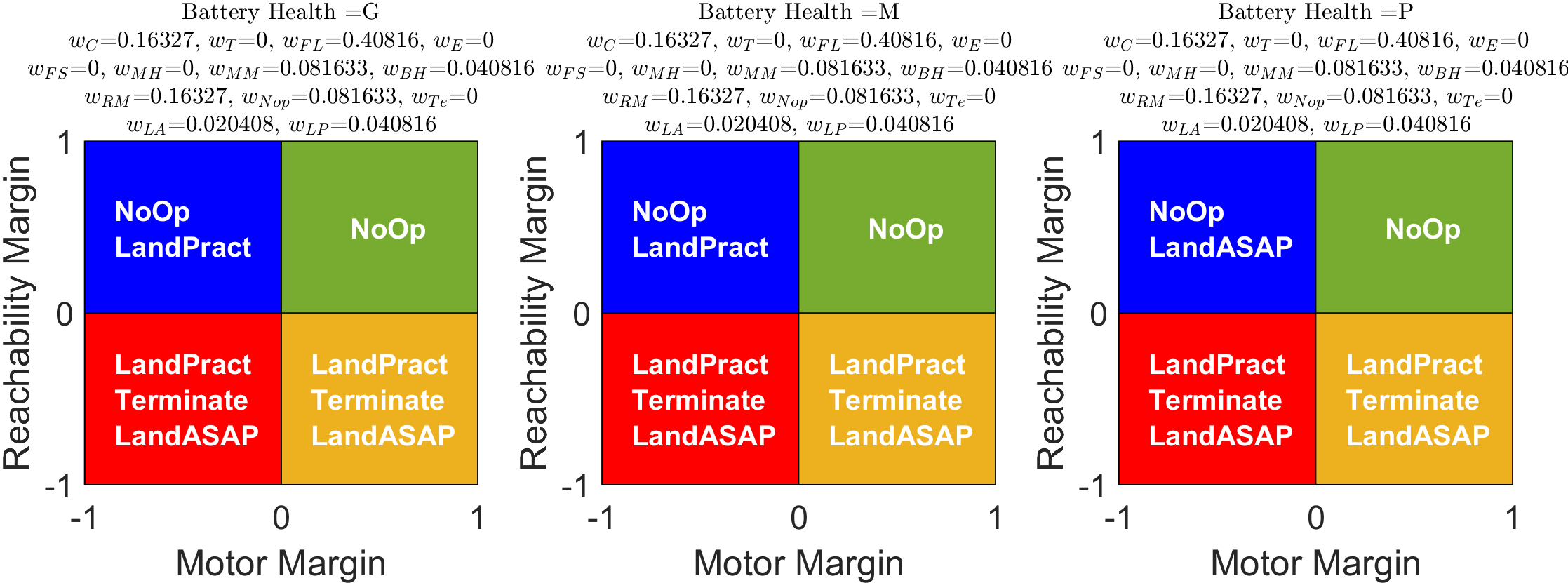}
    \caption{Reward sensitivity analysis, where the quadrant color signifies \textcolor{green}{$\blacksquare$}: Safe region, \textcolor{blue}{$\blacksquare$}: Hazardous region, \textcolor{yellow}{$\blacksquare$}: Critical region, \textcolor{red}{$\blacksquare$}: Hyper Critical region.   }
    \label{fig:RewSen}
\end{figure}

Weights $w_E$, $w_S$ and $w_A$ were tuned so the MDP optimal solution (Section~\ref{sec:mdpsolution}) matches our desired policy at specific key states as shown in Figure \ref{fig:RewSen}. The quadrants are defined based on $RM$ and $MM$ values. When battery health is poor, i.e. higher likelihood of failing the mission, the available emergency landing is $LandASAP$ in the hazardous region, as compared to scenarios where battery health is good or medium in which case the available emergency landing action is $LandPract$. The final weights selected were:
\begin{gather}
w_E(S) =\begin{cases}
0.163, &\text{$S=C$}\\
0.408, &\text{$S=FL$}\\
0, & \text{$S=\{T,E\}$}
\end{cases}, 
w_A(A) =\begin{cases}
0.082, &\text{$A = NoOp$}\\
0, &\text{$A = Terminate$}\\
0.02, & \text{$A=LandASAP$}\\
0.041, & \text{$A=LandPract$}
\end{cases} \\
w_S = [0,\; 0,\; 0.082,\; 0.041,\; 0.163]' 
\end{gather}

This method of tuning weights is similar to an inverse reinforcement learning (IRL) \cite{ng2000algorithms} approach in which reward function parameters are sought to recover an expert policy. One interpretation of this approach is that the MDP serves as a robust generalization mechanism, that is, an expert indicates the proper behavior in a few states, and the MDP generalizes this behavior to all states. The typical pitfalls of automated inverse reinforcement learning, such as the underspecified nature of the problem~ \cite{ng2000algorithms}, are avoided in this case because the manual tuning is overseen by an expert.

\subsubsection{Optimal MDP Solution}\label{sec:mdpsolution}
Given that infinite-horizon discounted MDPs can be solved in polynomial time (that is time polynomial in size of the state space and action space \cite{ye_simplex_2011}) and size of our problem's state and action space, iterative algorithms can easily find an optimal policy within tractable time. The CMA MDP optimal policy $\pi^*$ is calculated offline using value iteration with a decision epoch of $1Hz$. 

\begin{figure}[ht!]
    \centering
    \includegraphics[width=6in, height = 1.9in]{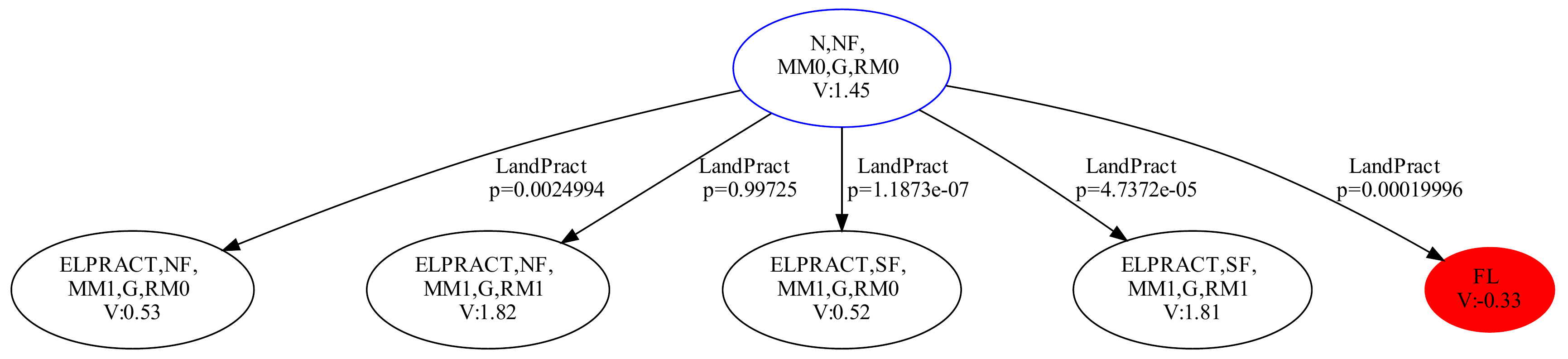}
    \caption{Optimal action that is executed from \textcolor{blue}{$S_t$} =$ \{N,$ $NF,$ $MM0,$ $G,$ $RM0\}$}
    \label{fig:oneStateOptim}
\end{figure}

 As an example, optimal action choice $LandPract$ from state \textcolor{blue}{$S_t$} = $\{N,$ $NF,$ $MM0,$ $G,$ $RM0\}$ is shown in Figure \ref{fig:oneStateOptim}. Both $MM<0$ and $RM<0$ indicate that the UAS is in critical condition. Sufficient energy is not available in the LiPo batteries to complete the nominal flight plan and at the same time, motors have either ceased, or RUL is very low to complete the flight plan. The best action for the UAS is to $LandPract$, as the battery health is good and this action selects an emergency landing flight plan from the database of much shorter duration than the original nominal flight plan.

\subsection{POMDP Model}\label{POMDPModel}
Many systems are modeled as fully observable processes, when in reality, some portion of the problem is partially observable. This portion of the work aims to address assumptions on the observability of the Markov decision process through developing a partially observable Markov decision process (POMDP) on which the performance of the optimal MDP policy and near-optimal POMDP policies can be evaluated. In short, the aim is to determine the value of partial observability in representing a component failure.
\par
In order to evaluate the trade-offs between a fully observable and partially observable approach to contingency management, a partially observable formulation of the CMA MDP is developed. This POMDP representation of the problem provides a benchmark on which both MDP and POMDP solution methods are evaluated.

A POMDP is defined by the 6-tuple $\langle S,A,p(s'|s,a),r(s,a),O,p(o|s',a) \rangle $, where $S$, $A$, $p(s'|s,a)$, $r(s,a)$ are defined in section \ref{sec:mdp_model}, $O$ is the finite set of system observations, and $p(o|s',a)$ is probability of obtaining a given observation conditioned on the state transitioned into and the action taken. 
 
\subsubsection{Observation Space}
The observation space $O$ of the POMDP is defined as follows:
\begin{equation}
    O = \textit{FS} \times \textit{MotorObs} \times \textit{RM}
\end{equation}
where, $ \textit{MotorObs} = \{MM0,MM1,JF\}$

Observations are defined as a tuple consisting of some subset of the true state, in this case a tuple consisting of flight status, reachability margin, and either motor margin or motor health according to the description below.

\subsubsection{Observation Probabilities}
Observation probabilities $p(o|s',a)$ are dependent on the true underlying state as defined below:
\begin{itemize}
    \item \textit{Flight Status:} True flight status state is deterministically observed at each time step
    \item \textit{Reachability Margin:} True reachability margin is observed with probability $P_o$, inaccurate reachability margin is observed with probability $1-P_o$.
    \item \textit{Motor Health:} Due to the ability of the IMM to rapidly resolve uncertainty around a jam fault (well under the timescale of the MDP), this fault is deterministically observable. When jam fault is observed, no motor margin is indicated as motor margin is always zero in jam fault. No observations of motor health are provided otherwise.
    \item \textit{Motor Margin:} When in the Spalling Fault state element, the true motor margin is observed with probability $P_o$, inaccurate motor margin is observed with probability $1-P_o$. The no fault state deterministically has motor margin 1 in simulation. Thus, when in no fault $MM1$ is observed with probability $P_o$ and $MM0$ is observed with probability $1-P_o$. These models account for uncertainty on both motor margin and fault type, where fault type is the primary source of uncertainty. In simulation, motor margin state $MM0$ is only possible if the motor is in Spalling Fault or Jam Fault states. Thus, by introducing uncertainty on the Motor Margin, uncertainty over the fault type is introduced where previously no errant motor margin indications were possible in the No Fault state.
    \item \textit{Battery Health:} No observations are provided of this state element. This state element is partially observable in the sense that it is informed by initial belief over states and transition dynamics only (which in this case do not allow for transition from one battery health to another).
\end{itemize}

\par
Ideally, the observation probability $P_o$ would be set using sensor accuracies from the literature. However since the MDP definition utilizes binary states, this would require translation of diagnosis error into binary observation probabilities. Furthermore, systems may have varying degrees of observability dependent on the specific equipment used. In lieu of one specific observation probability, we elect to evaluate the performance difference between POMDP and MDP policies at numerous observation probabilities. This sweep of observation probabilities provides more general guidance on what degrees of observability warrant a POMDP representation in the context of this Markov model.

\begin{figure}[ht!]
  \centering
    \includegraphics[width=1.0\textwidth]{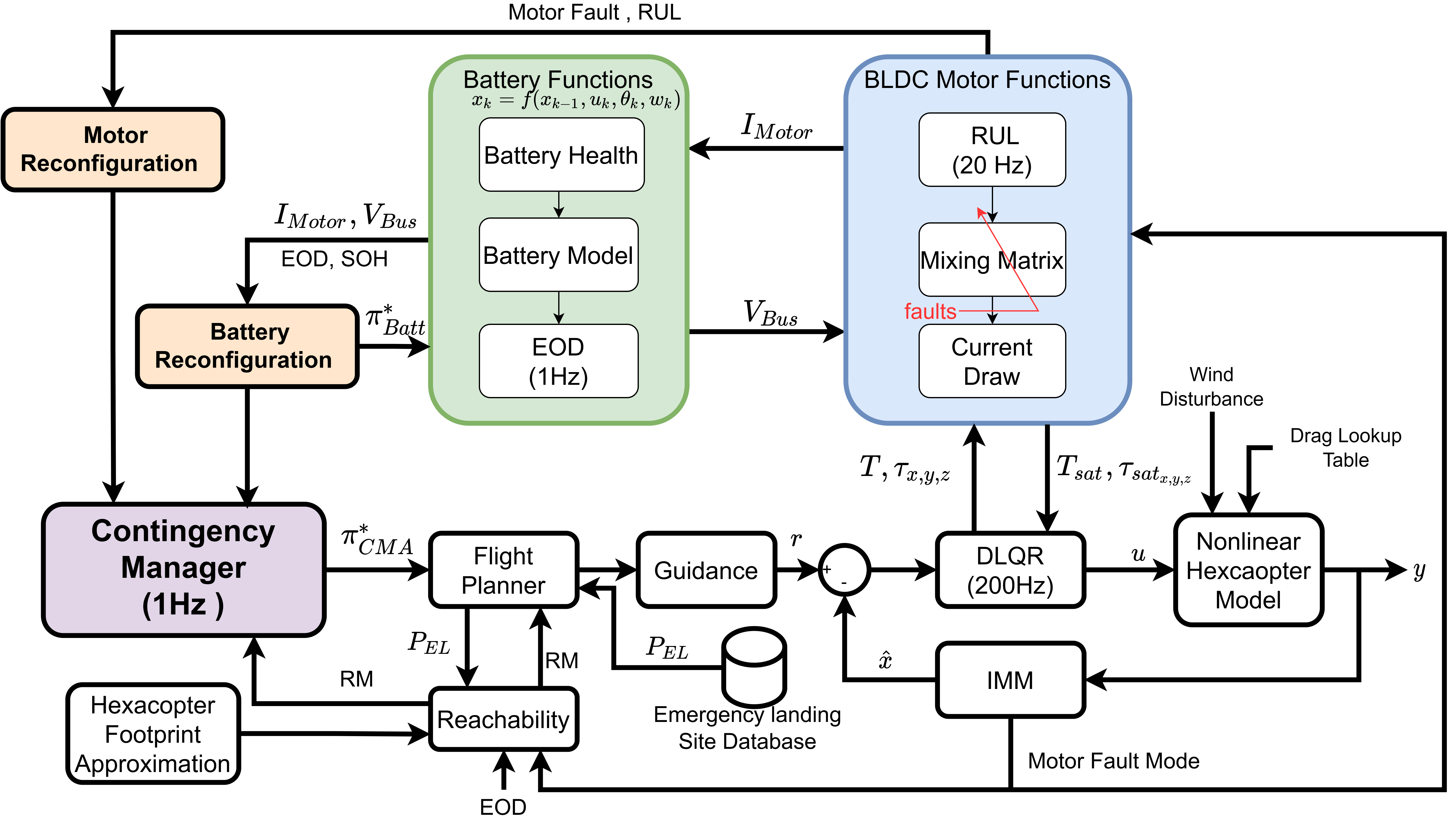}
    \caption{System diagram for simulating MDP and POMDP policies. }
    \label{fig:systemDiag}
\end{figure}

\section{Simulations}
\label{sec:Sim}
\subsection{System Architecture}

Before CMA can be deployed on an actual system, rigorous testing of CMA is required in a simulated environment with high fidelity models of system components. Performance of the CMA MDP and POMDP are evaluated in simulation based on experimentally validated models of a hexacopter, LiPo batteries and propulsion modules \cite{sharma2021prognostics}, \cite{sharma2019Hex}. Figure \ref{fig:systemDiag} shows the high fidelity simulation control loop developed for this paper. A hexacopter UAS was chosen to offer redundancy with minimal complexity in the propulsion system. The simulation is initiated with a package delivery flight plan input to the UAS guidance module. The flight planner receives information from the emergency landing site database, MDP contingency policy developed offline, and reachability module. As instructed by MDP actions the flight planner provides updates to the Guidance-Navigation-Control (GNC) hexacopter module.

The battery function block contains equivalent circuit resistance models of the LiPo battery, battery health identification module and End of Discharge calculator. The motor function module determines motor remaining useful life (RUL) once a spalling fault is detected.  Reconfiguration of the hexacopter motor mixing matrix is based on spalling fault properties, motor thrust saturation and motor current draw based on commanded thrust requirements.

To diagnose the presence of a rotor fault, a custom Interacting Multiple Model (IMM) filter was developed. The IMM is capable of delivering sufficient accuracy while remaining computationally tractable \cite{Shalom2002}. The reader is referred to \cite{Blom1988} for an in depth treatment of the IMM. The employed IMM maintains a probability distribution over seven regimes or modes of hexacopter operation. The first mode corresponds to nominal dynamics while the remaining six modes correspond to dynamics given a single failed hexacopter rotor.
For MDP case studies, full observability over rotor health is assumed and thus the maximum a posteriori (MAP) estimate is used to determine the hexacopter mode. Since the IMM as implemented is capable of resolving rotor failure uncertainty on a timescale well under that at which the POMDP calculations occur, the IMM diagnosis of the jam fault is taken to be fully observable in this context as well. 

These system modules are simulated at different rates to replicate actual hardware. The GNC module operates at $200Hz$. Battery end of discharge (EOD) calculations are updated at $1Hz$ since EOD does not change frequently.  Contingency policy execution is simulated at $1Hz$, matching the MDP formulation.  Failure in the simulation is defined as when the UAS is unstable or the available battery voltage is below the threshold voltage.     

\subsection{Policies and Policy Evaluation}\label{PolEval}
Solutions to the POMDP described above are provided using both a near optimal offline POMDP solution method, SARSOP \cite{kurniawati_sarsop_2008}, and through use of different measures of the belief state which allow an MDP policy (found using value iteration) to be evaluated on the POMDP. These measures of the belief take two different approaches to handling uncertainty in the state. 
\par
The previous observable MDP approximation ($Obs\;MDP$) treats the observation received at each time step as the true state and takes an optimal MDP action based on this presumed state. In practice, since the observation space is a subset of the state space, battery health is inferred from the initial belief, motor health is assumed to be no fault ($NF$) when motor margin is 1 ($MM1$), and motor health is assumed to be spalling fault ($SF$) when motor margin is 0 ($MM0$). The optimal MDP action from the value iteration policy is taken for this reconstructed state. This representation is meant to replicate the assumption that all observations are accurate. The second approximation is a maximum likelihood or maximum a posteriori (MAP) representation ($MAP\; MDP$). This representation takes the optimal MDP action for the state in the belief with the highest likelihood after a belief update.
\par
Observation probability is varied from perfect observations (correct subspace of the state or state features observed with probability $P=1$) to near uniform probability of observing either the correct state or incorrect state ($P=0.6$). Numerous simulations are run with each observation probability in order to assess metrics such as safety (terminate action avoided, landing successful) and efficiency (initial goal reached), in addition to reward.
\par 
For simulations executed on the high fidelity simulation architecture described above, the probability of a spalling fault occurring is $P(SF|NF)=0.5$. Note that this probability is higher than that used to find the optimal MDP policy and POMDP policies (one policy is found for each observability value) per Table \ref{table:MH}. Spalling faults lead to jam faults which may necessitate a modified trajectory (an emergency landing) or terminating the flight plan. Thus results represent a higher frequency of fault occurrence than expected under normal circumstances. This higher sampling is used in order to reduce the number of simulations necessary to evaluate policy performance. The MDP acting on the fully observable problem (state directly provided) $True\;MDP$ is run in order to provide an upper bound on performance. A baseline, state-independent \textit{NoOp} at every time step policy is also run for comparison. Note that this $Baseline\;NoOp$ policy does not depend on states or observations, thus it is shown across all observabilities. Likewise the fully observable MDP does not depend on observations, so it too is shown across all observabilities.

\section{Results}
\label{sec:Results}
Various case studies are conducted to determine the efficacy of CMA MDP for a UAS package delivery mission. Along with the case studies, we also conducted Monte Carlo (MC) simulations to determine the overall performance of MDP, POMDP policies and $Baseline\;NoOp$ policy. Parameters such as wind speed, direction, battery health, motor failure location, and state observation noise were varied during the MC simulations. Two metrics, namely Original Mission Completion Rate and Safety Rate, are used to assess the performance. These metrics provide an indication of policy efficiency and policy safety, respectively. The original flight plan completion rate represents a normalized value of completed trajectories provided by the flight planner at the start of the UAS mission. Safety rate is derived from compliment of failure rate of UAS simulated missions. These metrics are evaluated across three battery health initial conditions: Good, Medium, and Poor. 35 Monte Carlo simulations were performed at each observability and battery health for each policy.

\begin{figure*}[ht!]
    \centering
    \begin{subfigure}[t]{0.5\textwidth}
        \includegraphics[width = 1.02\textwidth]{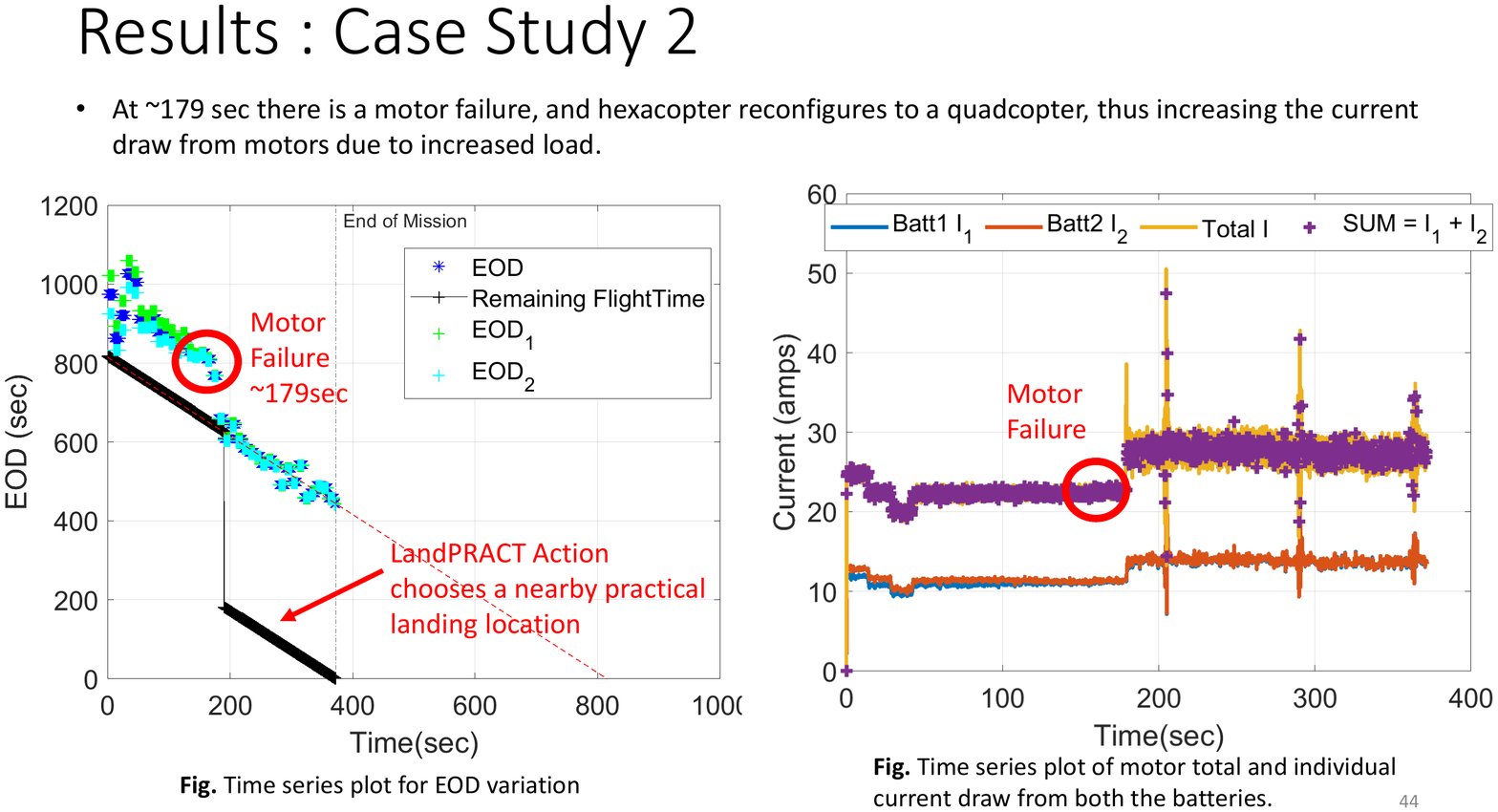}
        \caption{}
    	\label{fig:EODPolicy}
    \end{subfigure}%
    \begin{subfigure}[t]{0.5\textwidth}
        \includegraphics[width = 1.02\textwidth ]{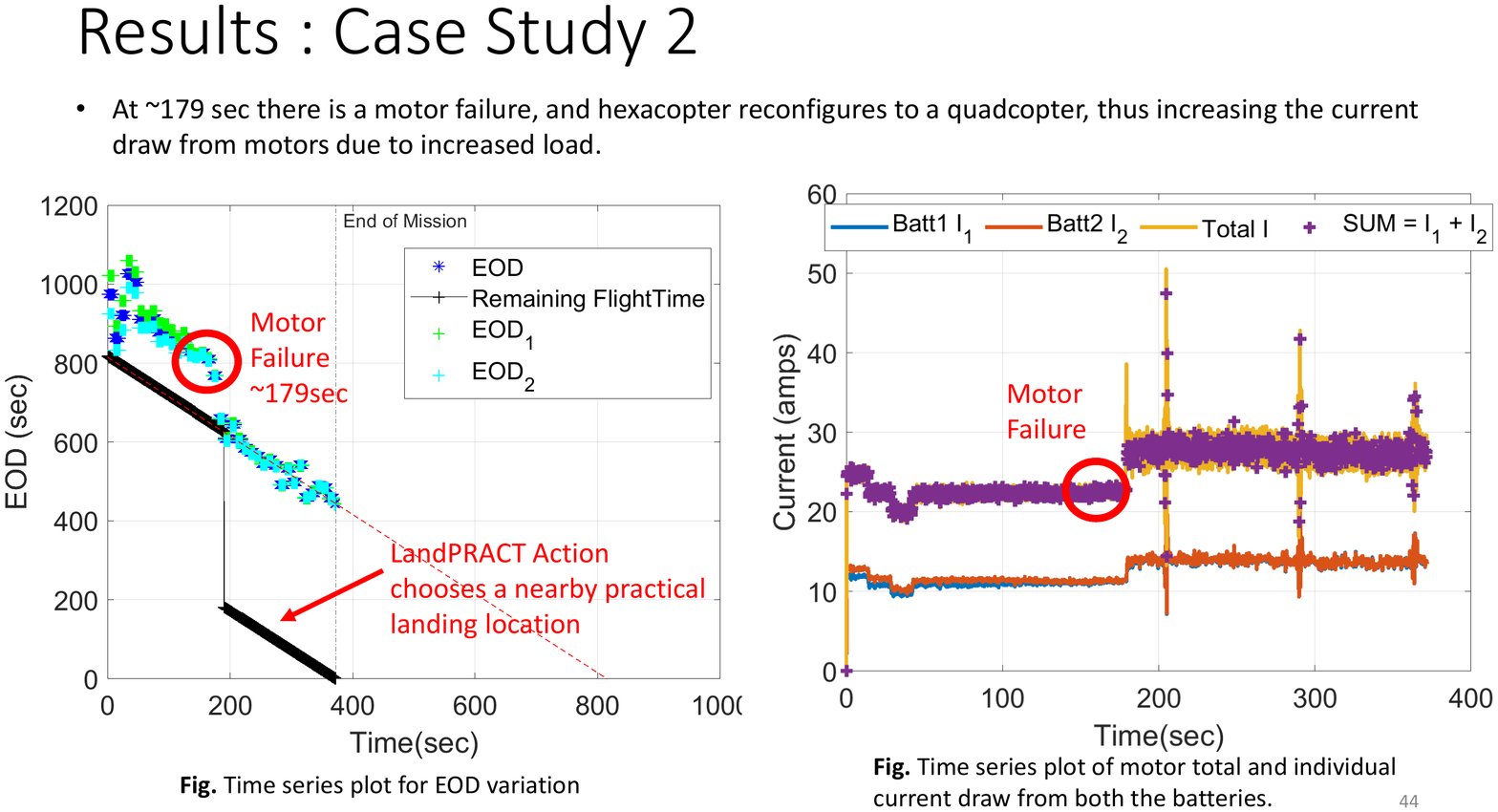}
        \caption{}
    	\label{fig:CurrPolicy}
    \end{subfigure}
    \caption{(a) EOD and remaining flight time variation with policy implementation, (b) Motor current consumption variation due to motor reconfiguration.}
    \label{fig:ECPolicy}
\end{figure*}

\begin{figure*}[ht!]
    \centering
    \begin{subfigure}[t]{0.5\textwidth}
        \centering
        \includegraphics[width = 1.02\textwidth]{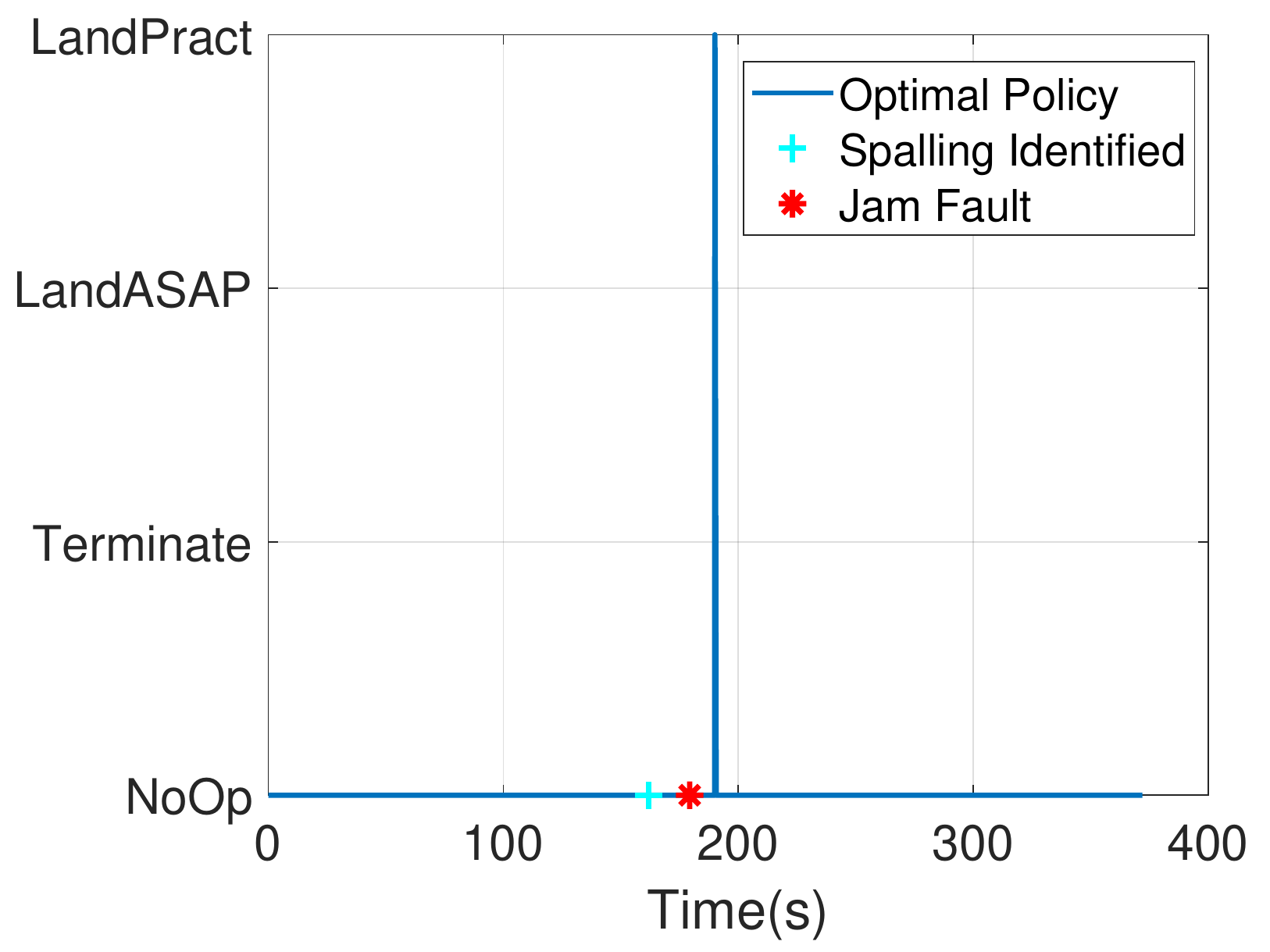} 
        \caption{}
    	\label{fig:OptPolicy}
    \end{subfigure}%
    \begin{subfigure}[t]{0.5\textwidth}
        \centering
        \includegraphics[width = 1.02\textwidth]{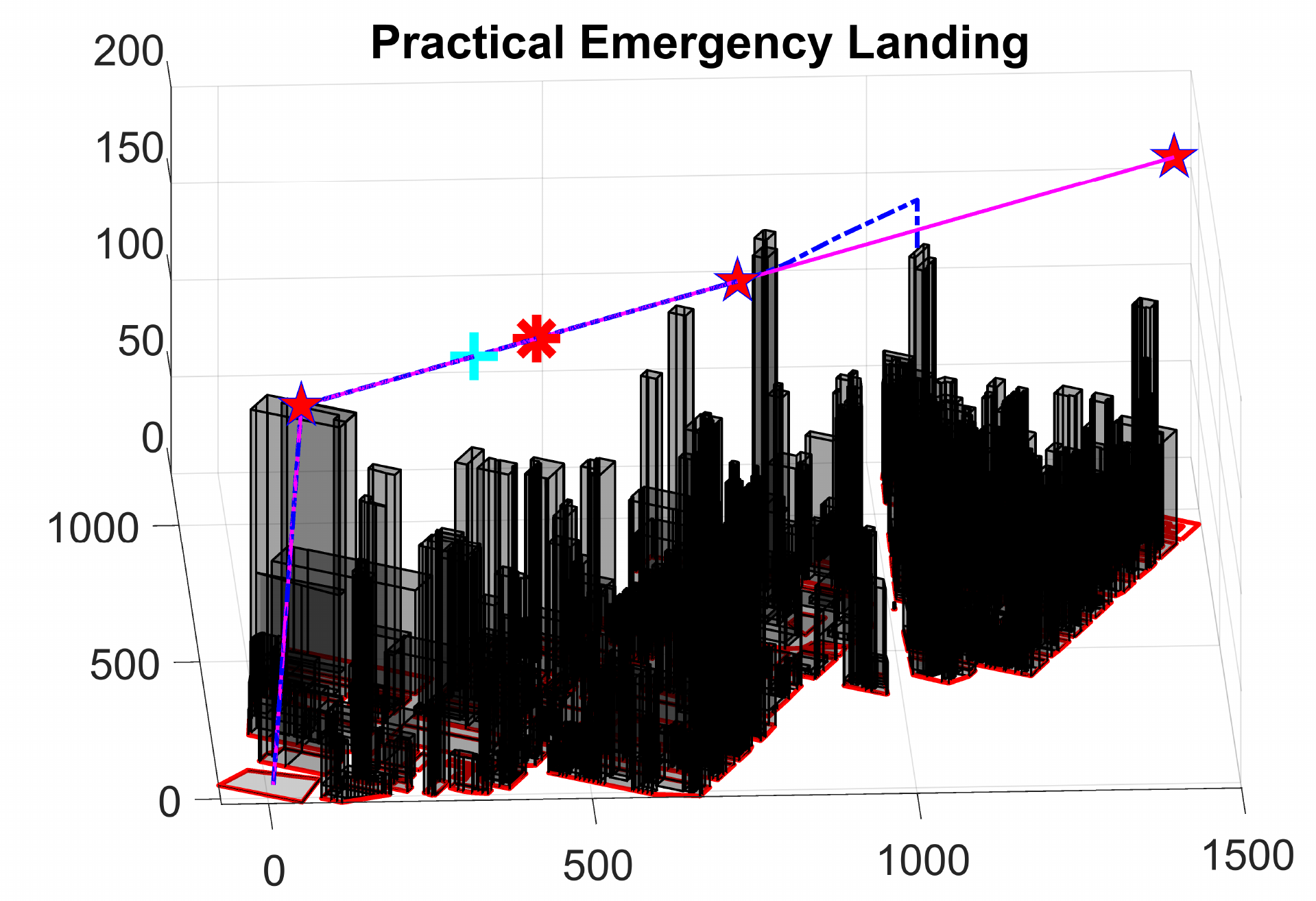}
        \caption{}
    	\label{fig:LandPracPolicy}
    \end{subfigure}
    \caption{(a) Time series plot of contingency management policy execution, (b) 3D plot showing execution of the CMA MDP policy in simulation where the blue trend is the \textcolor{blue}{Actual Flight Plan}, the magenta trend is the \textcolor{magenta}{Nominal Flight Plan}, and the red stars \color{red}$\bigstar$ are Checkpoints.}
    \label{fig:CMAPolicy}
\end{figure*}

\subsection{CMA Case Study}
This section highlights an example execution of the CMA MDP policy when the multicopter experiences motor degradation. The motor degradation eventually results in a jam fault with a cascading effect on EOD as shown in Figures \ref{fig:ECPolicy} and \ref{fig:CMAPolicy}. At $~161sec$ a spalling fault is detected and prognosis is performed to determine motor RUL. The CMA MDP continues to recommend a $NoOp$ action as shown in Figure \ref{fig:OptPolicy}. The motor fails at $~179 sec$ with failure detected by the IMM. This leads the hexacopter to reconfigure into a quadrotor configuration. With this reconfiguration higher thrust is demanded from the four motors, resulting in an increase in current draw from the battery shown in Figure \ref{fig:CurrPolicy}. This sudden increase in current drawn from the battery reduces the EOD values to an unexpectedly low level as seen in Figure \ref{fig:EODPolicy}, and the resulting EOD value is insufficient to complete the flight plan. The CMA MDP identifies this condition and recommends the $LandPract$ action shown in Figure \ref{fig:OptPolicy}. The $LandPract$ action triggers the search of an appropriate pre-flight calculated emergency landing trajectory with maximum reachability margin from the nearest checkpoint shown as a red star in Figure \ref{fig:LandPracPolicy}. An emergency landing trajectory is selected and sent to GNC. While the UAS is executing the emergency landing flight plan, the CMA-MDP is still active but its available action set is reduced to $A={NoOp, Terminate, LandASAP}$. In this case study the UAS successfully landed at the selected emergency landing site.  


\subsection{Original Flight Plan Completion Rate vs Safety}

\par
\begin{figure*}[h!]
        \centering
        \includegraphics[width=0.6\textwidth]{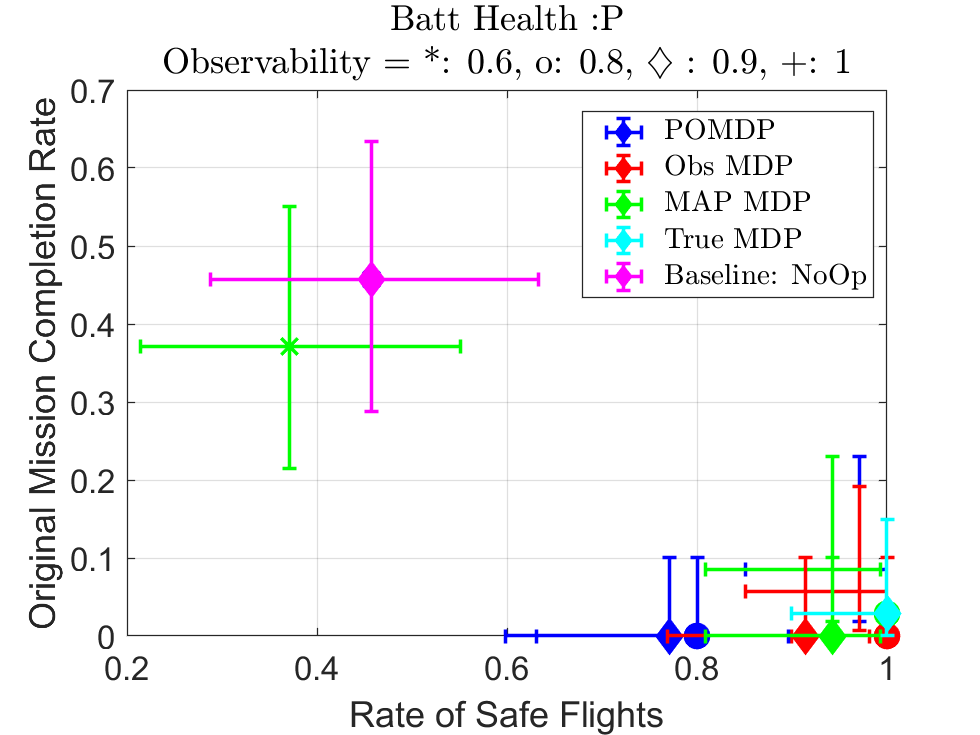}
        \caption{Original mission completion rate versus the rate of safe flights for poor battery health with 95\% confidence bounds. Note that the POMDP policy at $P_{obs}=0.6$ is obscured by the $Baseline\;NoOp$ policy.}
        \label{fig:FvsSP}
\end{figure*}

\begin{figure*}[h!]
        \centering
        \includegraphics[width=0.6\textwidth]{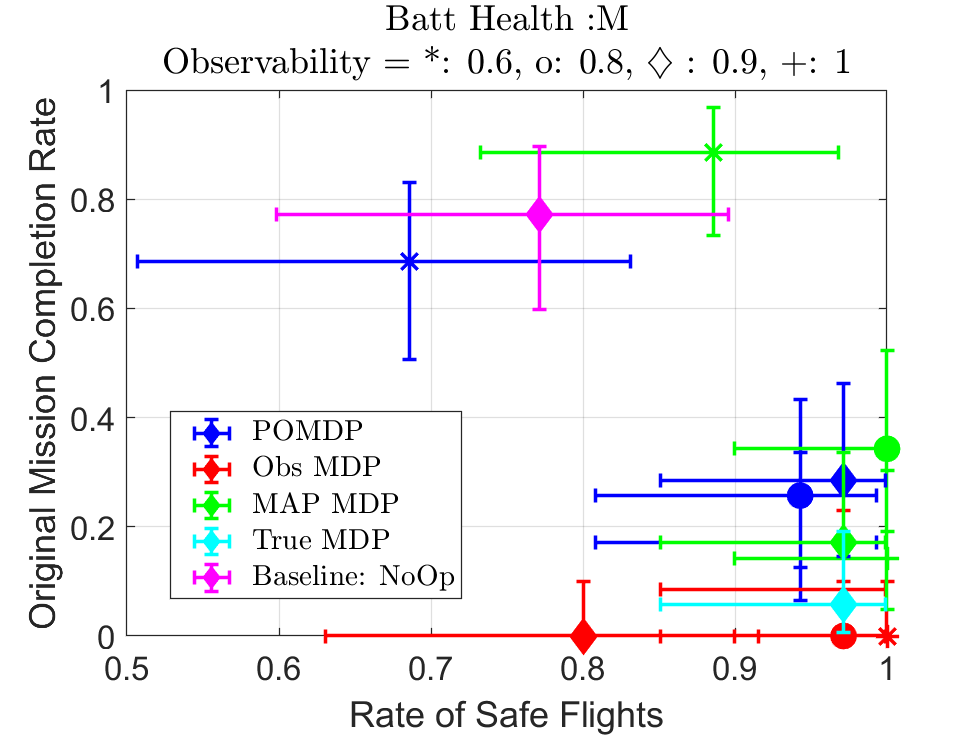}
        \caption{Original mission completion rate versus the rate of safe flights for medium battery health with 95\% confidence bounds.}
        \label{fig:FvsSM}
\end{figure*}

\begin{figure*}[h!]
        \centering
        \includegraphics[width=0.6\textwidth]{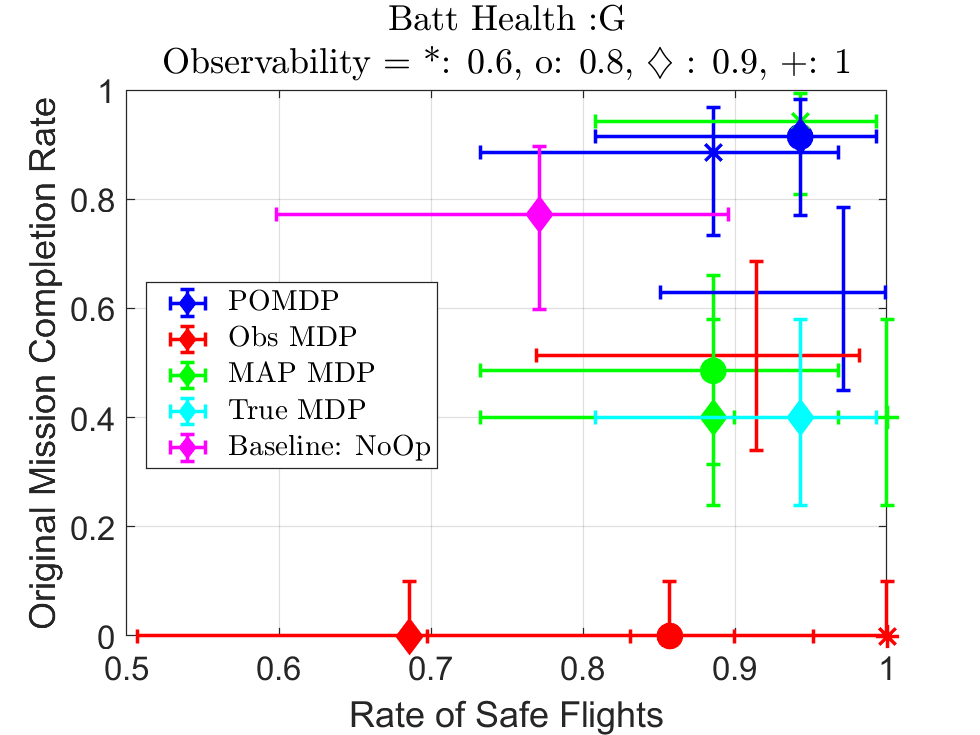}
        \caption{Original mission completion rate versus the rate of safe flights for good battery health with 95\% confidence bounds.}
        \label{fig:FvsSG}
\end{figure*}

 Across all battery health metrics, the previous observation MDP, $Obs\;MDP$, performs poorly in partial observability, completing nearly zero flights on average while maintaining varying degrees of safety (Figure \ref{fig:FvsSP}-\ref{fig:FvsSG}) In the case where several false fault observations are received in a row, there are two explanations for this behavior. The first is that the policy elects to land very early in the flight in response to these observations, in which case we would see safe, but inefficient flights. The second is that these observations lead the policy to select several contingency actions in rapid succession. This selection in several cases led the lower-level controller to destabilize or the state to be incorrectly labeled \textit{jam fault} when there was in fact no jam fault, resulting in a mission failure, which in reality would be restricted by a lower level controller preventing such cases. Were controller stability or misdiagnosis not an issue, we would expect to observe strictly safe and inefficient (not completing the original mission) behavior from this policy under partial observability, due to its treating any fault observation as accurate and taking a contingency action.

\par With respect to the other policies, the following trends are observed. With poor battery health (Figure \ref{fig:FvsSP}), we observe conservative policy behavior for all policies, except $Baseline\;NoOp$, at all observabilities greater than $P_{obs} = 0.6$. The $MAP\; MDP$ either outperforms or is comparable to the $POMDP$ policy in these cases, achieving a higher number of safe flights for the same completion rate, although results for both are within the 95\% confidence interval due to the limited number of trials. 
With poor battery health and observabilities greater than $P_{obs} = 0.6$, all policies perform comparably to the $True\;MDP$ which has access to the full state.  
\par At observability 0.6, the $POMDP$ and $MAP\; MDP$ policies are substantially less conservative, completing the intended mission roughly 40\% of the time, while completing flights safely only 40\% of the time. This performance is comparable to that of the $Baseline\;NoOp$ policy. This change in behavior can likely be attributed to a greater dependence on transition dynamics over observations, since observations provide very little information in this case. Since the transition matrix model assumes very low probabilities of rotor fault, the belief will remain concentrated on no fault states and continue along the planned trajectory. 

With medium battery health (Figure \ref{fig:FvsSM}), we observe similar trends (note the axis change). In this case, policies are more efficient and safe, likely reflecting the increased margin of safety due to a healthier battery. At higher observabilities ($P_{obs} = 1$,  $P_{obs} = 0.9$, $P_{obs} =0.8$), the $MAP\; MDP$ and $POMDP$ perform comparably to each other and the $True\;MDP$ policy (within the 95\% confidence bounds). The same aggressive behavior is observed at the lowest observability, $P_{obs} = 0.6$.

With good battery health (Figure \ref{fig:FvsSG}), a change in trends is observed. At $P_{obs}=0.6$, the $MAP\; MDP$ and $POMDP$ perform comparably and substantially better than the $True\;MDP$ policy in terms of mission completion. The $POMDP$ performs better than the $MAP\; MDP$ policy at the other observabilities ($P_{obs} = 1$,  $P_{obs} = 0.9$, $P_{obs} =0.8$) in terms of mission completion, while the $MAP\; MDP$ performs much more comparably to the $True\;MDP$ policy. Interestingly, the partially observable cases perform better than the full observability $POMDP$ and $MAP\; MDP$ policies. This is likely a continuation of the same aggressive behavior observed in prior plots under poor observability, $P_{obs} =0.6$. However, under good battery conditions the UAS is much more likely to remain safe and complete missions, unlike in the previous cases. It is worth noting that the $POMDP$ policy outperforms the $Baseline\;NoOp$ policy under good battery conditions. 

\subsection{CMA Performance based on Battery Health}
\begin{figure*}[ht!]
    \centering
     \begin{subfigure}[t]{0.475\textwidth}
        \centering
        \includegraphics[width=1.03\textwidth]{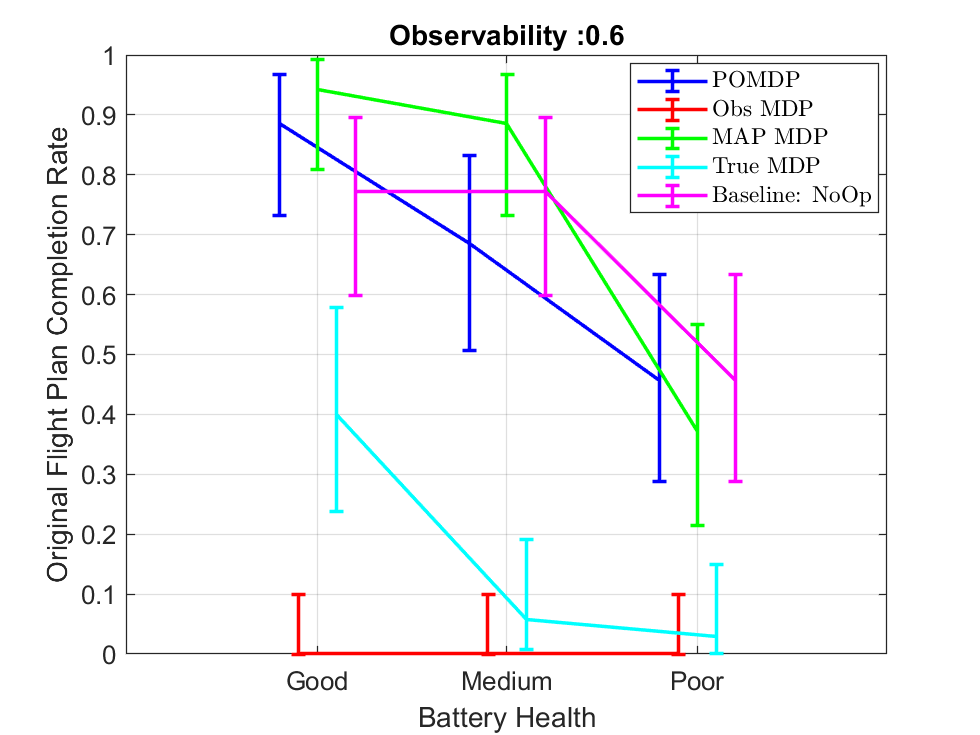}
        \caption{60\% Partial State Observability}
        \label{fig:Obs0.6MissRate}
    \end{subfigure}
  \begin{subfigure}[t]{0.475\textwidth}
        \centering
        \includegraphics[width=1.03\textwidth]{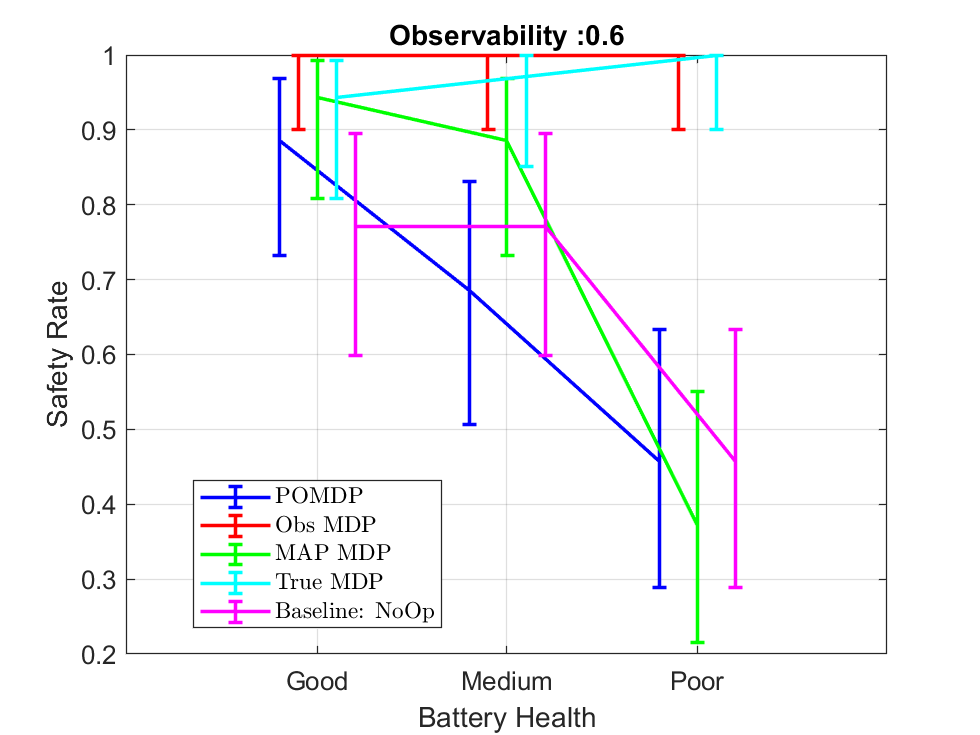}
        \caption{60\% Partial State Observability}
        \label{fig:Obs0.6Safety}
    \end{subfigure}
     \begin{subfigure}[t]{0.475\textwidth}
        \centering
        \includegraphics[width=1.03\textwidth]{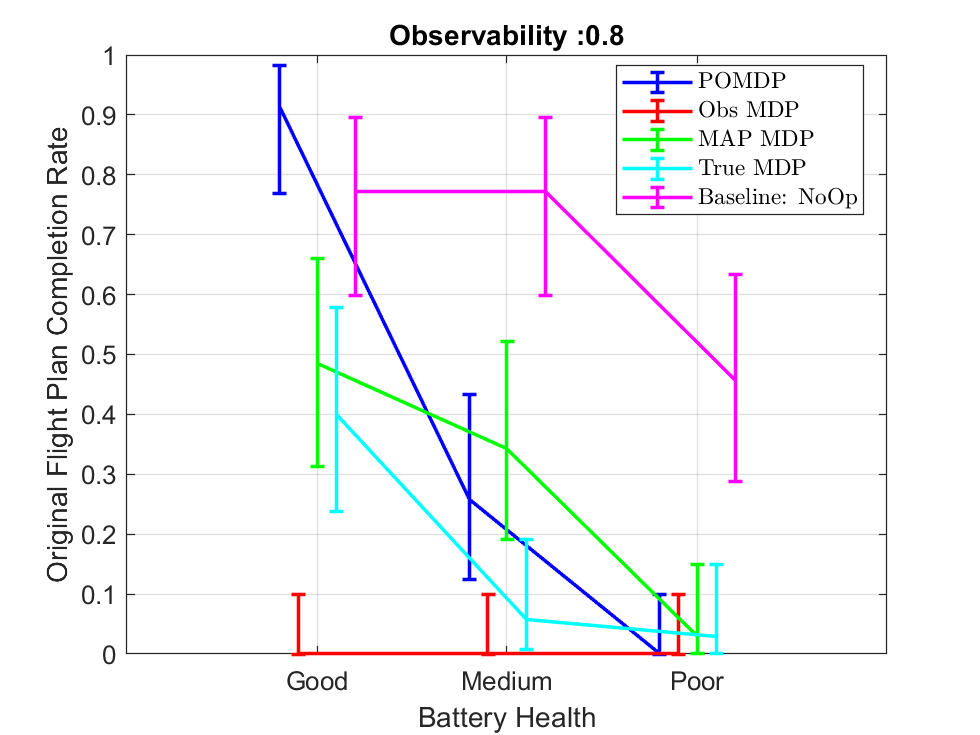}
        \caption{80\% State Observability}
        \label{fig:Obs0.8MissRate}
    \end{subfigure}
       \begin{subfigure}[t]{0.475\textwidth}
        \centering
        \includegraphics[width=1.03\textwidth]{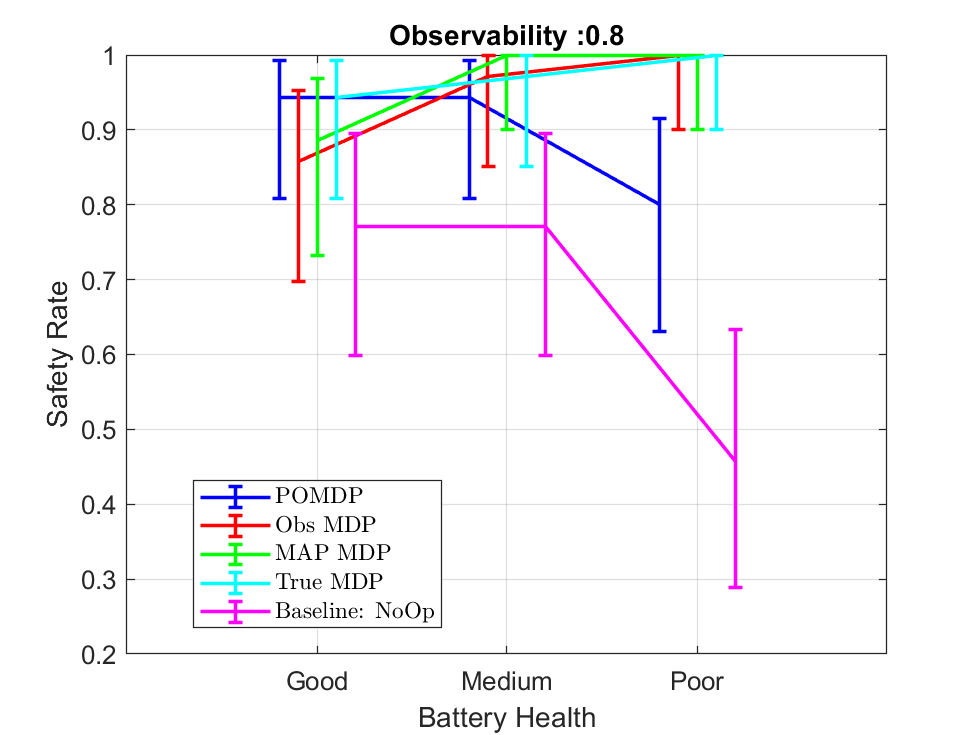}
        \caption{80\% Partial State Observability}
        \label{fig:Obs0.8Safety}
    \end{subfigure}
    \begin{subfigure}[t]{0.475\textwidth}
        \centering
        \includegraphics[width=1.03\textwidth]{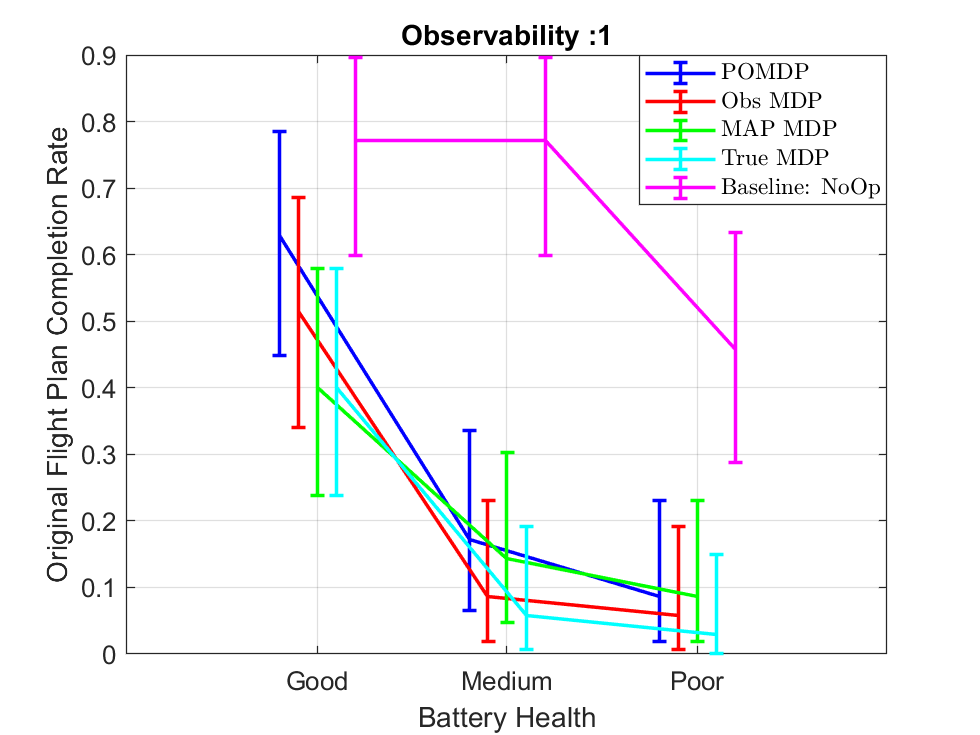}
        \caption{Full State Observability}
        \label{fig:Obs1MissRate}
    \end{subfigure}
    \begin{subfigure}[t]{0.475\textwidth}
        \centering
        \includegraphics[width=1.03\textwidth]{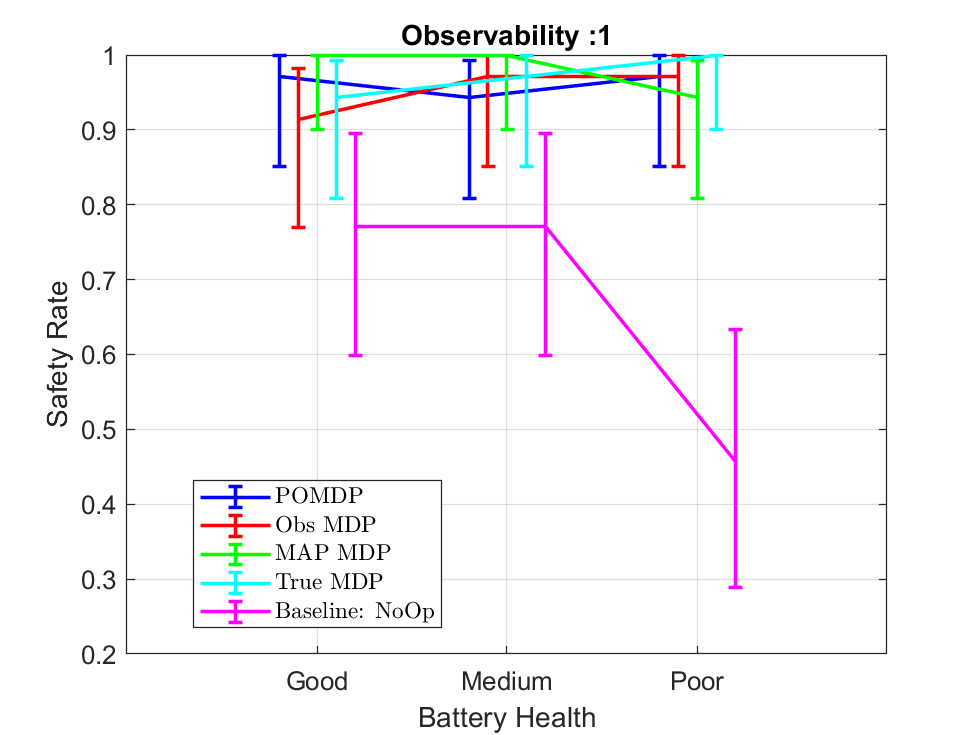}
        \caption{Full State Observability}
        \label{fig:Obs1Safety}
    \end{subfigure}
  \caption{Safety rate and original flight plan completion rate over different battery health conditions in high fidelity simulation for contingency policies. The $Baseline\;NoOp$ and $True\;MDP$ policies are not affected by observability. }
    \label{fig:ObsMissSafety}
\end{figure*}

This section presents results analyzing the performance of the contingency management policies under different battery health and observability conditions shown in Figure \ref{fig:ObsMissSafety}. 

 The Original Flight Plan completion rate of all the contingency policies with poor state observability $0.6$, $0.8$ and full state observability are shown in Figures \ref{fig:Obs0.6MissRate}, \ref{fig:Obs0.8MissRate} and  \ref{fig:Obs1MissRate}. Considering full state observability case Figure \ref{fig:Obs1MissRate}, all the policies have a drop in performance as the battery health degrades. As expected the $POMDP,$ $Obs \;MDP$, $MAP \;MDP$ are in the vicinity of the $True\;MDP$ policy under full observability. All of these policies have relatively low mission completion performance compared with the $Baseline\;NoOp$ policy. This is because the contingency policies aggressively execute emergency landing in scenarios where $MM <0$, i.e. a single motor is about to fail or has already failed and $RM > 0$, i.e. it has sufficient energy to complete the flight plan. Similar aggressive safety behaviour is exhibited in medium and poor battery health condition also. In poor state observability scenarios shown in Figure \ref{fig:Obs0.6MissRate} and \ref{fig:Obs0.8MissRate} the $True\;MDP$ and $Baseline\; NoOp$ are shown only as a reference for comparison, since they are not affected by the state observability. $Obs\; MDP$ has the lowest mission completion rate with very noisy state observation and executes emergency landing early on in the flight plan. Further as the state observability degrades there is an increase in the performance of $POMDP$ and $MAP\; MDP$ policies. This is attributed to the aggressive behaviour towards completing the flight plan with less concern about safety.

From analyzing the safety rate for different battery health conditions over different state observability the $True\;MDP$ and $Baseline\;NoOp$ policies represent the upper and lower bounds, respectively. At full observability in Figure \ref{fig:Obs1Safety}, the CMA MDP and POMDP policies converge to $True\;MDP$ and exhibit higher safety then the $Baseline\;NoOp$ policy. With poor state observability the CMA MDP and POMDP policies converge to $Baseline\;NoOp$ policy. The $Obs\;MDP$ policy as shown in Figure \ref{fig:Obs0.6Safety} might appear the safest but that is because it never completes any mission in poor state observability scenarios and lands  early in the flight plan. In summary, the conservative behaviour of the contingency policies results in higher safety rates but lower mission completion rates.

\subsection{Transition Matrix Simulations}
\label{sec:tmat_res}

In addition to high fidelity simulations presented above, 'transition matrix simulations' were executed. These simulations apply the transition matrix used to calculate the optimal policy as the evaluation simulator. This is in contrast to the high fidelity simulator, which is closer to a real world deployment of the policy and uses different state transition dynamics. Both simulators use the same observation matrices. These transition matrix simulations allow for isolation of sources of error in our application of the abstracted POMDP model solutions to the high fidelity simulation. The relevant potential sources of error are: poor abstraction of the real world dynamics to POMDP state transition dynamics and poor translation of desired outcomes to objectives. In the experiments previously described, there is potential for both dynamics and objective mismatches. In contrast, the following simulations control for differences in dynamics by evaluating the performance of the above policies on the transition matrix simulator, where the dynamics of the solution and problem necessarily match. 5000 Monte Carlo evaluations of the policies were completed using the transition matrix simulator with 100 maximum simulation steps. 
\par
\begin{figure*}[h!]
    \centering
    \begin{subfigure}[t]{0.475\textwidth}
        \centering
        \includegraphics[width=1.0\textwidth]{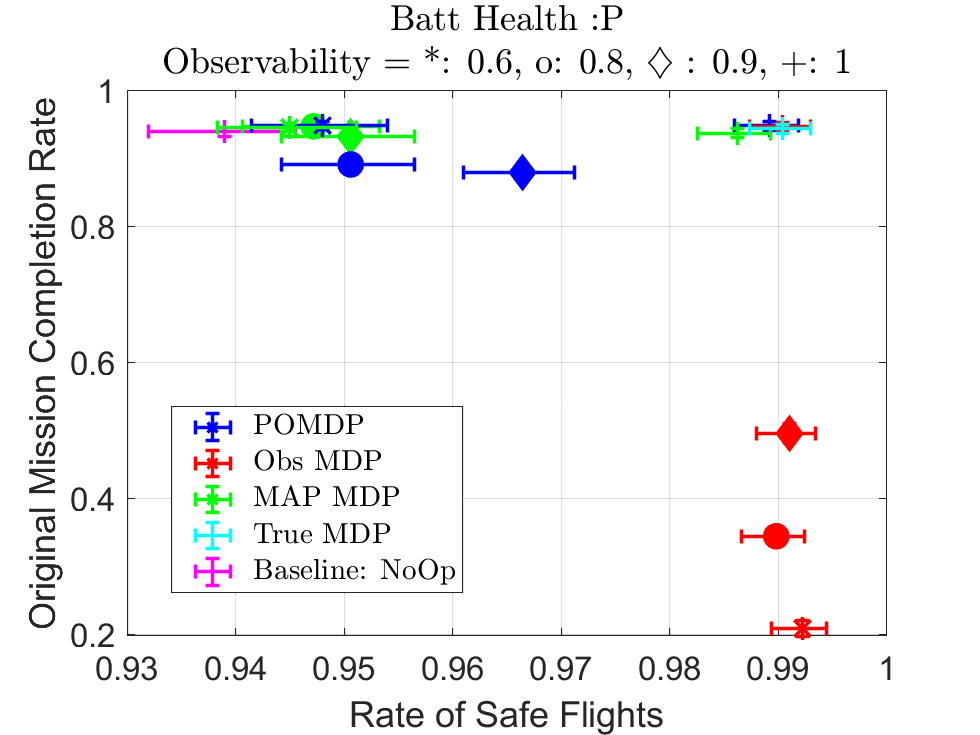}
        \caption{}
        \label{fig:tmat1}
    \end{subfigure}
    \begin{subfigure}[t]{0.475\textwidth}
        \centering
        \includegraphics[width=1.0\textwidth]{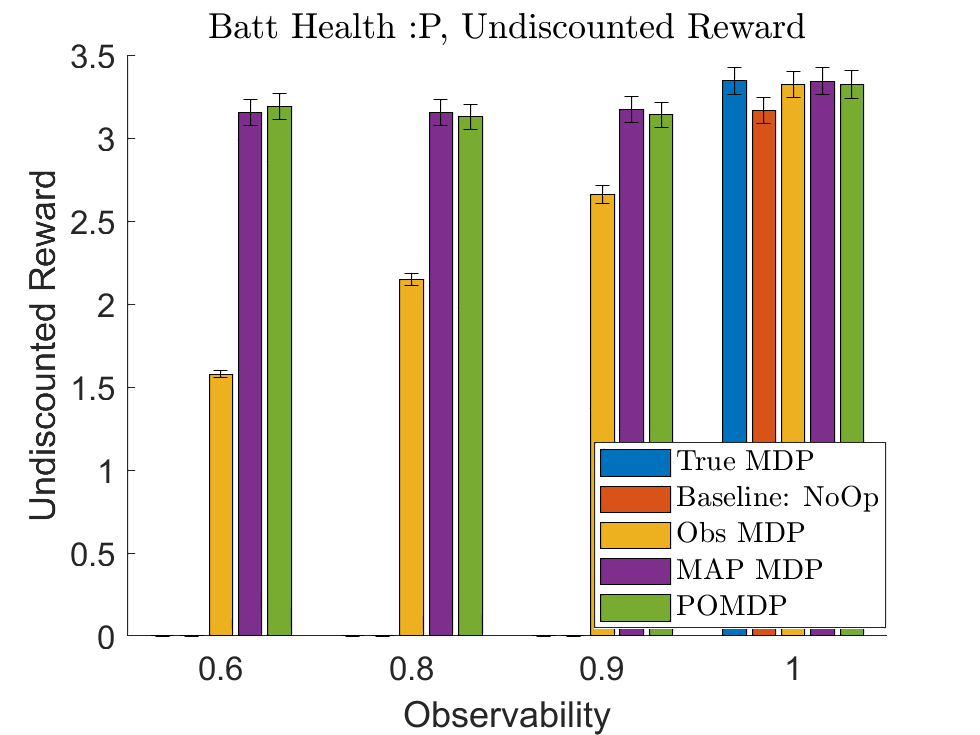}
        \caption{}
        \label{fig:tmat2}
    \end{subfigure}
    \caption{Transition matrix simulations for poor battery health. (a) Original mission completion rate versus the rate of safe flights with 95\% confidence bounds. Note the axis are at different scales. (b) Average (undiscounted) reward for the same simulations, with $2\sigma$ standard error of the mean bounds. Note that the $True\;MDP$ and $Baseline\;NoOp$ policy do not depend on the observability and are only shown at full observability. }
    \label{fig:tmat}
\end{figure*}

Figure \ref{fig:tmat1} shows mission completion rates versus safe flight rates, using the same metrics as in Figures \ref{fig:FvsSP}-\ref{fig:FvsSG}, while controlling for dynamics mismatch. In these simulations, $POMDP$ and $MAP\; MDP$ achieve similar performance in the transition matrix simulations, while the $Obs\;MDP$ policy is safe, but inefficient. In cases where $POMDP$ and $MAP\; MDP$ are not comparable, the $POMDP$ policy performs more conservatively, completing fewer missions while remaining more safe. In this case, where no differences in dynamics are present, we observe potential objective mismatch, as the POMDP policy behaves more conservatively than other policies.
\par 
Figure \ref{fig:tmat2} shows the average cumulative reward for each policy and degree of observability, controlling for both dynamics and objective mismatch. Thus this plot indicates if the policies are near-optimal with respect to the abstracted dynamics and the objectives as defined in the state transition and reward functions, respectively. The $MAP\; MDP$ and $POMDP$ policies achieve comparable performance to each other across observabilites in terms of reward. $POMDP$ and $MAP\; MDP$ perform comparably to the $True\;MDP$ and the $Baseline\;NoOp$ policies at full observability and substantially better than the $Obs\;MDP$ policy at decreased observabilities. $MAP\; MDP$ and $POMDP$ policies are near-optimal with respect to the reward of the abstracted POMDP formulation. 
\par Note that comparable performance in terms of reward does not directly translate to equivalent performance in terms of safety or efficiency. This is a result of the reward function design, which was not designed to explicitly account for the completion and safety metrics discussed above. Given this near-optimal reward performance shown in Figure \ref{fig:tmat2} and the higher mission completion and safe flight rates shown in Figure \ref{fig:tmat1}, the most likely contributor to error in this work is the abstracted state transition dynamics.

\section{Discussion}
\label{sec:Discussion}

\subsection{Policy Evaluation using High Fidelity Simulations}
\par
The above results reveal UAS missions have a low completion rate under poor battery conditions for both the $POMDP$ and $MAP\; MDP$ in Poor and Medium Battery Conditions. This safety preserving behavior of CMA does lead to less failures than the $Baseline\;NoOp$ policy. However, one would expect an optimal policy to be less conservative (achieve nominal mission completions closer to that of the $Baseline\;NoOp$ policy) while still achieving a lower failure rate.
\par 
It is of note that the high fidelity simulator generated several state transitions which were not supported by the MDP transition matrix.  At times the system appears to transition directly from no fault to a jam fault due to miscategorized system behavior. Unrepresented or inaccurately modeled transitions may contribute to the lower completion and safety rates observed.  
\par
Unexpected transitions in these underlying states may explain the higher completion rates observed with higher partial observability states. In simulations with low $P_o$, errant states would be attributed to an incorrect observation, where the deterministic observation model with $P_o=1$ cannot account for unexpected transitions as observation noise.
Unexpected state transitions bolster the argument for accounting for noisy observations. Given that it appears the state itself indicated by the high fidelity simulator is subject to some noise, accounting for this noise in the model is all the more important. 

These trends aside, in all but the battery health good initial condition, the $POMDP$ policy and the $MAP\; MDP$ representation achieve similar performance, as is expected given the transition matrix simulations.
Although the $Baseline\;NoOp$ policy generally has the worst performance in terms of safety rate, this policy performs as well as or better than the others in terms of the original flight plan completion rate. This is especially interesting when noting that the $True\;MDP$ which has access to the state (directly, without uncertainty) performs worse than the $Baseline\;NoOp$ policy in terms of nominal missions completed. We would hope to see MDP completion rate performance closer to that of the $Baseline\;NoOp$ while still maintaining a much lower failure rate, though these metrics inevitably compete. Because the primary goal of a CMA policy is safety of the UAS and not mission completion, these results are not surprising.
\par
As noted in \ref{PolEval},a higher probability of spalling fault occurrence is used in the high fidelity simulations than in the transition matrix. With a lower probability of spalling (as would be expected in a real world system), an increase in performance is expected, both in terms of safety and efficiency. Thus, the results presented reflect a worse-case scenario of sorts, where failure is much more likely than it ever should be in a real world setting.
Due to the relatively small sample sizes used in high fidelity testing, the number of simulations with possible spalling faults and the number of simulations where spalling faults manifested is not consistent across policies and battery conditions.

\subsection{Policy Evaluation using Transition Matrix Simulations}

\par As noted in Section \ref{sec:tmat_res}, the $POMDP$ and $MAP\; MDP$ policies achieve similar performance with respect to reward, safety, and efficiency. One possible explanation for this similar performance is that the belief over states remains concentrated on one state over time. In this case we would expect a policy based on the MAP estimate of the state and a policy based the full belief to achieve similar performance. To quantify the degree to which the belief  distribution remains concentrated, the minimum MAP state probability, $P_{MinMax}$, is introduced in eq. \ref{eq:MinMaxBel}. This value is the minimum MAP state probability across the time horizon of a simulation. The highest probability in the belief is found for each time step. From this set of MAP probabilities, the minimum is taken. This provides an indication of how concentrated the belief was over the course of a given simulation.
\begin{equation}\label{eq:MinMaxBel}
    P_{MinMax}(i) = \min_{t \in h}\{\max_{s \in b_t}b_{t}(s) | t \in h\}
\end{equation}
where $h$ is the horizon of the simulation, $t$ is the time step, $s$ is the state, $b$ are the probabilities assigned to states in the belief, and $i$ is the number of the simulation.

\par Across 60000 transition matrix simulations of fixed horizon 100 steps (as above) of the $MAP\; MDP$ policy, 5000 with each initial battery state and observability, the mean of this metric is $0.991$ with standard deviation $\sigma = 0.031$, while the minimum is $0.5073$. Thus, most simulations maintain a low uncertainty belief. This can be attributed to the probability of transition to fault states encoded in the transition matrix. As an example, consider deterministic transitions. These would lead to a concentrated belief regardless of observation, as an observation in conflict with the state which is deterministically transitioned to will simply be attributed to the probability of incorrect observation. Thus, probabilities which are nearly deterministic also lead to concentrated belief.
\par
Given these results, the $MAP\; MDP$ policy may be the best policy for this context. Finding this policy only has the computational complexity of finding the value iteration solution to the MDP. This policy benefits from maintaining a belief over states and is more robust to noisy observations for this reason.


\section{Conclusion}
\label{sec:Conclusion}
\par
In this work we have proposed a risk-aware Contingency Management Autonomy (CMA) built on the powerful framework of Markov Decision Processes. Experimentally validated hexacopter UAS, propulsion unit, and battery performance and degradation models are incorporated into MDP and POMDP formulations.  The effectiveness of the contingency management policies is evaluated on a high fidelity simulator, which more closely represents a real world application, and also on a transition matrix simulator, which allows for investigation of model accuracy.

The policies are evaluated over various component degradation scenarios and state observabilities. Two metrics, namely original flight plan completion rate and safety rate, are used to analyse the performance of the different policies. In our approach, active information gathering and long time uncertainty are not present, as a result $MAP \; MDP$ is favourable. This policy has reduced computational complexity when compared to $POMDP$. The $MAP \; MDP$ policy also outperforms a naive MDP policy implementation which assumes that all observations are accurate. We conclude that a maximum a posterior approximation, or similar method that uses an MDP policy while reasoning over beliefs, may be the best compromise between solution fidelity and computational complexity in formulations where there are observation uncertainties and nearly deterministic state transitions. 

\par
In future work we aim to refine the state transition model so that it accurately reflects the transition probabilities of the simulator. Success of CMA is currently evaluated in simulations. To further validate our results the next step would be to conduct hardware-based experiments aimed at determining the efficacy of CMA in practice.




\section*{Acknowledgments}
Funding for this work was provided by industry sponsors through the Center for Unmanned Aircraft Systems NSF IUCRC, grant no. 1650468.
\bibliography{elanding}

\begin{thebibliography}{43}
\newcommand{\enquote}[1]{``#1''}
\providecommand{\natexlab}[1]{#1}
\providecommand{\url}[1]{\texttt{#1}}
\providecommand{\urlprefix}{URL }
\expandafter\ifx\csname urlstyle\endcsname\relax
  \providecommand{\doi}[1]{\discretionary{}{}{}https://doi.org/#1}\else
  \providecommand{\doi}[1]{\discretionary{}{}{}\urlstyle{rm}\url{https://doi.org/#1}}\fi

\bibitem[{Osborne et~al.(2019)Osborne, Lantair, Shafiq, Zhao, Robu, Flynn, and
  Perry}]{osborne2019uas}
Osborne, M., Lantair, J., Shafiq, Z., Zhao, X., Robu, V., Flynn, D., and Perry,
  J., \enquote{UAS operators safety and reliability survey: Emerging
  technologies towards the certification of autonomous UAS,} \emph{2019 4th
  International Conference on System Reliability and Safety (ICSRS)}, IEEE,
  2019, pp. 203--212.

\bibitem[{Ochoa and Atkins(2022)}]{ochoa2022urban}
Ochoa, C.~A., and Atkins, E.~M., \enquote{Urban Metric Maps for Small Unmanned
  Aircraft Systems Motion Planning,} \emph{Journal of Aerospace Information
  Systems}, Vol.~19, No.~1, 2022, pp. 37--52.

\bibitem[{Di~Donato and Atkins(2017)}]{di2017evaluating}
Di~Donato, P.~F., and Atkins, E.~M., \enquote{Evaluating risk to people and
  property for aircraft emergency landing planning,} \emph{Journal of Aerospace
  Information Systems}, Vol.~14, No.~5, 2017, pp. 259--278.

\bibitem[{E.{Balaban} and J.J.{Alonso}(2013)}]{balaban2013modeling}
E.{Balaban}, and J.J.{Alonso}, \enquote{A modeling framework for prognostic
  decision making and its application to uav mission planning,} \emph{Annual
  Conference of The Prognostics and Health Management Society}, 2013, pp.
  1--12.

\bibitem[{{Balaban} et~al.(2013){Balaban}, S.{Narasimhan}, M.{Daigle},
  I.{Roychoudhury}, A.{Sweet}, C.{Bond}, and
  G.{Gorospe}}]{balaban2013development}
{Balaban}, E., S.{Narasimhan}, M.{Daigle}, I.{Roychoudhury}, A.{Sweet},
  C.{Bond}, and G.{Gorospe}, \enquote{Development of a mobile robot test
  platform and methods for validation of prognostics-enabled decision making
  algorithms,} \emph{International Journal of Prognostics and Health
  Management}, Vol.~4, No.~1, 2013, p.~87.

\bibitem[{{Schacht-Rodríguez} et~al.(2019){Schacht-Rodríguez}, {Ponsart},
  {García-Beltrán}, {Astorga-Zaragoza}, and {Theilliol}}]{schacht2019}
{Schacht-Rodríguez}, R., {Ponsart}, J.~C., {García-Beltrán}, C.~D.,
  {Astorga-Zaragoza}, C.~M., and {Theilliol}, D., \enquote{Mission planning
  strategy for multirotor UAV based on flight endurance estimation*,}
  \emph{International Conference on Unmanned Aircraft Systems (ICUAS)}, 2019,
  pp. 778--786.
\newblock \urlprefix\url{https://doi.org/10.1109/ICUAS.2019.8798292}.

\bibitem[{{Tang} et~al.(2008){Tang}, {Kacprzynski}, {Goebel}, {Saxena}, {Saha},
  and {Vachtsevanos}}]{tang2008}
{Tang}, L., {Kacprzynski}, G.~J., {Goebel}, K., {Saxena}, A., {Saha}, B., and
  {Vachtsevanos}, G., \enquote{Prognostics-enhanced Automated Contingency
  Management for advanced autonomous systems,} \emph{International Conference
  on Prognostics and Health Management}, 2008, pp. 1--9.
\newblock \urlprefix\url{https:/doi/org/10.1109/PHM.2008.4711448}.

\bibitem[{{Tang} et~al.(2010){Tang}, {Kacprzynski}, {Goebel}, and
  {Vachtsevanos}}]{tang2010}
{Tang}, L., {Kacprzynski}, G.~J., {Goebel}, K., and {Vachtsevanos}, G.,
  \enquote{Case studies for prognostics-enhanced Automated Contingency
  Management for aircraft systems,} \emph{IEEE Aerospace Conference}, 2010, pp.
  1--11.
\newblock \urlprefix\url{https://doi.org/10.1109/AERO.2010.5446844}.

\bibitem[{{Zhang} et~al.(2014){Zhang}, L.{Tang}, J.{Decastro}, M.{Roemer}, and
  {Goebel}}]{zhang2014}
{Zhang}, B., L.{Tang}, J.{Decastro}, M.{Roemer}, and {Goebel}, K.,
  \enquote{Autonomous Vehicle Battery State-of-Charge Prognostics Enhanced
  Mission Planning,} \emph{International Journal of Prognostics and Health
  Management}, Vol.~5, No.~8, 2014.

\bibitem[{Schacht-Rodr{\'\i}guez et~al.(2018)Schacht-Rodr{\'\i}guez, Ponsart,
  Garcia-Beltran, and Astorga-Zaragoza}]{schacht2018prognosis}
Schacht-Rodr{\'\i}guez, R., Ponsart, J.-C., Garcia-Beltran, C.~D., and
  Astorga-Zaragoza, C.~M., \enquote{Prognosis \& health management for the
  prediction of uav flight endurance,} \emph{IFAC-PapersOnLine}, Vol.~51,
  No.~24, 2018, pp. 983--990.

\bibitem[{He et~al.(2013)He, Williard, Chen, and Pecht}]{he2013state}
He, W., Williard, N., Chen, C., and Pecht, M., \enquote{State of charge
  estimation for electric vehicle batteries using unscented kalman filtering,}
  \emph{Microelectronics Reliability}, Vol.~53, No.~6, 2013, pp. 840--847.

\bibitem[{Daigle and Goebel(2010)}]{daigle2010improving}
Daigle, M.~J., and Goebel, K.~F., \enquote{Improving computational efficiency
  of prediction in model-based prognostics using the unscented transform,}
  \emph{Annual Conference of the Prognostics and Health}, 2010.

\bibitem[{Dalal et~al.(2011)Dalal, Ma, and He}]{dalal2011lithium}
Dalal, M., Ma, J., and He, D., \enquote{Lithium-ion battery life prognostic
  health management system using particle filtering framework,}
  \emph{Proceedings of the Institution of Mechanical Engineers, Part O: Journal
  of Risk and Reliability}, Vol. 225, No.~1, 2011, pp. 81--90.

\bibitem[{Obeid et~al.(2020)Obeid, Tariq, and
  Mukhopadhyay}]{obeid2020supervised}
Obeid, A., Tariq, U., and Mukhopadhyay, S., \enquote{Supervised learning for
  early and accurate battery terminal voltage collapse detection,} \emph{IET
  Circuits, Devices \& Systems}, Vol.~14, No.~3, 2020, pp. 347--356.

\bibitem[{Wu et~al.(2016)Wu, Fu, and Guan}]{wu2016review}
Wu, L., Fu, X., and Guan, Y., \enquote{Review of the remaining useful life
  prognostics of vehicle lithium-ion batteries using data-driven
  methodologies,} \emph{Applied Sciences}, Vol.~6, No.~6, 2016, p. 166.

\bibitem[{Liu et~al.(2013)Liu, Pang, Zhou, Peng, and
  Pecht}]{liu2013prognostics}
Liu, D., Pang, J., Zhou, J., Peng, Y., and Pecht, M., \enquote{Prognostics for
  state of health estimation of lithium-ion batteries based on combination
  Gaussian process functional regression,} \emph{Microelectronics Reliability},
  Vol.~53, No.~6, 2013, pp. 832--839.

\bibitem[{Sharma and Atkins(2021)}]{sharma2021prognostics}
Sharma, P., and Atkins, E., \enquote{Prognostics-Informed Battery
  Reconfiguration in a Multi-Battery Small UAS Energy System,} \emph{2021
  International Conference on Unmanned Aircraft Systems (ICUAS)}, IEEE, 2021,
  pp. 423--432.

\bibitem[{Brown et~al.(2015)Brown, Coffey, Harvey, and
  Thayer}]{brown2015characterization}
Brown, J.~M., Coffey, J.~A., Harvey, D., and Thayer, J.~M.,
  \enquote{Characterization and prognosis of multirotor failures,}
  \emph{Structural Health Monitoring and Damage Detection, Volume 7}, Springer,
  2015, pp. 157--173.

\bibitem[{Zhang et~al.(2013)Zhang, Chamseddine, Rabbath, Gordon, Su, Rakheja,
  Fulford, Apkarian, and Gosselin}]{zhang2013development}
Zhang, Y., Chamseddine, A., Rabbath, C.~A., Gordon, B.~W., Su, C.-Y., Rakheja,
  S., Fulford, C., Apkarian, J., and Gosselin, P., \enquote{Development of
  advanced FDD and FTC techniques with application to an unmanned quadrotor
  helicopter testbed,} \emph{Journal of the Franklin Institute}, Vol. 350,
  No.~9, 2013, pp. 2396--2422.

\bibitem[{Dydek et~al.(2012)Dydek, Annaswamy, and
  Lavretsky}]{dydek2012adaptive}
Dydek, Z.~T., Annaswamy, A.~M., and Lavretsky, E., \enquote{Adaptive control of
  quadrotor UAVs: A design trade study with flight evaluations,} \emph{IEEE
  Transactions on control systems technology}, Vol.~21, No.~4, 2012, pp.
  1400--1406.

\bibitem[{Mueller and D'Andrea(2014)}]{mueller2014stability}
Mueller, M.~W., and D'Andrea, R., \enquote{Stability and control of a
  quadrocopter despite the complete loss of one, two, or three propellers,}
  \emph{2014 IEEE international conference on robotics and automation (ICRA)},
  IEEE, 2014, pp. 45--52.

\bibitem[{Kim et~al.(2021)Kim, Sharma, Atkins, Neogi, Dill, and
  Young}]{kim2021assured}
Kim, J., Sharma, P., Atkins, E., Neogi, N., Dill, E., and Young, S.,
  \enquote{Assured Contingency Landing Management for Advanced Air Mobility,}
  \emph{2021 IEEE/AIAA 40th Digital Avionics Systems Conference (DASC)}, IEEE,
  2021, pp. 1--12.

\bibitem[{Ten~Harmsel et~al.(2017)Ten~Harmsel, Olson, and
  Atkins}]{ten2017emergency}
Ten~Harmsel, A.~J., Olson, I.~J., and Atkins, E.~M., \enquote{Emergency flight
  planning for an energy-constrained multicopter,} \emph{Journal of Intelligent
  \& Robotic Systems}, Vol.~85, No.~1, 2017, pp. 145--165.

\bibitem[{Sunberg et~al.(2017)Sunberg, Ho, and
  Kochenderfer}]{zsunberg_value_of_inferring}
Sunberg, Z., Ho, C., and Kochenderfer, M., \enquote{The Value of Inferring the
  Internal State of Traffic Participants for Autonomous Freeway Driving,} ,
  2017.
\newblock \doi{10.48550/ARXIV.1702.00858},
  \urlprefix\url{https://arxiv.org/abs/1702.00858}.

\bibitem[{Kurniawati(2022)}]{kurniawati2022partially}
Kurniawati, H., \enquote{Partially Observable Markov Decision Processes and
  Robotics,} \emph{Annual Review of Control, Robotics, and Autonomous Systems},
  Vol.~5, 2022.

\bibitem[{Kaelbling et~al.(1998)Kaelbling, Littman, and
  Cassandra}]{kaelbling1998planning}
Kaelbling, L.~P., Littman, M.~L., and Cassandra, A.~R., \enquote{Planning and
  acting in partially observable stochastic domains,} \emph{Artificial
  intelligence}, Vol. 101, No. 1-2, 1998, pp. 99--134.

\bibitem[{Kochenderfer et~al.(2012)Kochenderfer, Holland, and
  Chryssanthacopoulos}]{kochenderfer2012next}
Kochenderfer, M.~J., Holland, J.~E., and Chryssanthacopoulos, J.~P.,
  \enquote{Next-generation airborne collision avoidance system,} Tech. rep.,
  Massachusetts Institute of Technology-Lincoln Laboratory Lexington United
  States, 2012.

\bibitem[{Goel et~al.(2000)Goel, Dedeoglu, Roumeliotis, and
  Sukhatme}]{Goel2000}
Goel, P., Dedeoglu, G., Roumeliotis, S.~I., and Sukhatme, G.~S., \enquote{Fault
  detection and identification in a mobile robot using multiple model
  estimation and neural network,} \emph{Proceedings - IEEE International
  Conference on Robotics and Automation}, Vol.~3, 2000, pp. 2302--2309.
\newblock \doi{10.1109/ROBOT.2000.846370}.

\bibitem[{Mehra et~al.(1998)Mehra, Rago, and Seereeram}]{Mehra1998}
Mehra, R., Rago, C., and Seereeram, S., \enquote{Autonomous failure detection,
  identification and fault-tolerant estimation with aerospace applications,}
  \emph{IEEE Aerospace Conference Proceedings}, Vol.~2, 1998, pp. 133--138.
\newblock \doi{10.1109/AERO.1998.687904}.

\bibitem[{Zhang and Li(1998)}]{Zhang1998}
Zhang, Y., and Li, X.~R., \enquote{Detection and diagnosis of sensor and
  actuator failures using IMM estimator,} \emph{IEEE Transactions on Aerospace
  and Electronic Systems}, Vol.~34, 1998, pp. 1293--1313.
\newblock \doi{10.1109/7.722715}.

\bibitem[{Kochenderfer et~al.(2022)Kochenderfer, Wheeler, and
  Wray}]{kochenderfer2022algorithms}
Kochenderfer, M.~J., Wheeler, T.~A., and Wray, K.~H., \emph{Algorithms for
  decision making}, Mit Press, 2022.

\bibitem[{Littman et~al.(1995)Littman, Cassandra, and
  Kaelbling}]{littman1995learning}
Littman, M.~L., Cassandra, A.~R., and Kaelbling, L.~P., \enquote{Learning
  policies for partially observable environments: Scaling up,} \emph{Machine
  Learning Proceedings 1995}, Elsevier, 1995, pp. 362--370.

\bibitem[{Kurniawati et~al.(2008)Kurniawati, Hsu, and
  Lee}]{kurniawati_sarsop_2008}
Kurniawati, H., Hsu, D., and Lee, W.~S., \enquote{{SARSOP}: {Efficient}
  point-based {POMDP} planning by approximating optimally reachable belief
  spaces,} \emph{In {Proc}. {Robotics}: {Science} and {Systems}}, 2008.

\bibitem[{Kim and Atkins(2022)}]{kim2022airspace}
Kim, J., and Atkins, E., \enquote{Airspace Geofencing and Flight Planning for
  Low-Altitude, Urban, Small Unmanned Aircraft Systems,} \emph{Applied
  Sciences}, Vol.~12, No.~2, 2022, p. 576.

\bibitem[{M.L.{Putterman}(2005)}]{Martin2005}
M.L.{Putterman}, \emph{Markov Decision Processes: Discrete Stochastic Dynamic
  Programming}, John Wiley \& Sons, 2005.
\newblock \urlprefix\url{https://doi.org/10.1002/9780470316887}.

\bibitem[{Qiu et~al.(2020)Qiu, Wu, Xu, Qiu, and Xue}]{qiu2020}
Qiu, C., Wu, X., Xu, C., Qiu, X., and Xue, Z., \enquote{An Approximate
  Estimation Approach of Fault Size for Spalled Ball Bearing in Induction Motor
  by Tracking Multiple Vibration Frequencies in Current,} \emph{Sensors},
  Vol.~20, No.~6, 2020.
\newblock \doi{10.3390/s20061631},
  \urlprefix\url{https://www.mdpi.com/1424-8220/20/6/1631}.

\bibitem[{Zhang et~al.(2009)Zhang, Sconyers, Patrick, and
  Vachtsevanos}]{zhang2009}
Zhang, B., Sconyers, C., Patrick, R., and Vachtsevanos, G., \enquote{A
  Multi-Fault Modeling Approach for Fault Diagnosis and Failure Prognosis of
  Engineering Systems,} \emph{Annual Conference of the Prognostics and Health
  Management Society}, , No. 1465, 2009.

\bibitem[{Atkins et~al.(2006)Atkins, Portillo, and
  Strube}]{atkins2006emergency}
Atkins, E.~M., Portillo, I.~A., and Strube, M.~J., \enquote{Emergency flight
  planning applied to total loss of thrust,} \emph{Journal of aircraft},
  Vol.~43, No.~4, 2006, pp. 1205--1216.

\bibitem[{Ng et~al.(2000)Ng, Russell et~al.}]{ng2000algorithms}
Ng, A.~Y., Russell, S., et~al., \enquote{Algorithms for Inverse Reinforcement
  Learning.} \emph{ICML}, Vol.~1, 2000, p.~2.

\bibitem[{Ye(2011)}]{ye_simplex_2011}
Ye, Y., \enquote{The {Simplex} and {Policy}-{Iteration} {Methods} {Are}
  {Strongly} {Polynomial} for the {Markov} {Decision} {Problem} with a {Fixed}
  {Discount} {Rate},} \emph{Mathematics of Operations Research}, Vol.~36,
  No.~4, 2011, pp. 593--603.
\newblock \doi{10.1287/moor.1110.0516},
  \urlprefix\url{https://pubsonline.informs.org/doi/10.1287/moor.1110.0516},
  publisher: INFORMS.

\bibitem[{Sharma and Atkins(2019)}]{sharma2019Hex}
Sharma, P., and Atkins, E., \enquote{Experimental Investigation of Tractor and
  Pusher Hexacopter Performance,} \emph{Journal of Aircraft}, Vol.~56, No.~5,
  2019, pp. 1920--1934.
\newblock \doi{10.2514/1.C035319},
  \urlprefix\url{https://doi.org/10.2514/1.C035319}.

\bibitem[{Bar-Shalom et~al.(2002)Bar-Shalom, Kirubarajan, and Li}]{Shalom2002}
Bar-Shalom, Y., Kirubarajan, T., and Li, X.-R., \emph{Estimation with
  Applications to Tracking and Navigation}, John Wiley \& Sons, 2002, Chap.~11,
  pp. 453--457.
\newblock \urlprefix\url{https://doi.org/10.1002/0471221279}.

\bibitem[{Blom and Bar-Shalom(Aug. 1988)}]{Blom1988}
Blom, H. A.~P., and Bar-Shalom, Y., \enquote{The interacting multiple model
  algorithm for systems with Markovian switching coefficients,} \emph{IEEE
  Transactions on Automatic Control}, Vol.~33, No.~8, Aug. 1988, pp. 780--783.
\newblock \doi{10.1109/9.1299}.

\end{thebibliography}

\end{document}